\documentclass{article}

\usepackage{microtype}
\usepackage{graphicx}
\usepackage{booktabs} % for professional tables
\usepackage{lscape}

\usepackage{hyperref}
\usepackage{subfig}    % For subfloat

\usepackage[accepted]{icml2023}

\usepackage{amsmath}
\usepackage{amssymb}
\usepackage{mathtools}
\usepackage{amsthm}

\usepackage[capitalize,noabbrev]{cleveref}

\usepackage{enumitem}
\usepackage{multirow}

\newcommand{\eg}{e.\,g.\,, }
\newcommand{\ie}{i.\,e.\,, }

\newcommand{\wrt}{w.\,r.\,t.}

\newcommand{\mX}{\textbf{X}}
\newcommand{\mY}{\textbf{Y}}
\newcommand{\mK}{\textbf{K}}

\newcommand{\mpr}{\textbf{p}}
\newcommand{\my}{\textbf{y}}
\newcommand{\mW}{\textbf{W}}
\newcommand{\mI}{\textbf{I}}

\newcommand{\adv}{Adv}
\newcommand{\bln}{Baln}
\newcommand{\norm}{FaiReg}
\newcommand{\normbln}{FaiRegH}
\newcommand{\expected}[1]{\mathbb{E}[#1]}
\newcommand{\variance}[1]{\mathbb{V}ar[#1]}
\newcommand{\cexpected}[2]{\mathbb{E}_{#1}[#2]}

\theoremstyle{plain}
\newtheorem{theorem}{Theorem}[section]

\theoremstyle{definition}
\newtheorem{definition}[theorem]{Definition}

\theoremstyle{remark}

\usepackage[textsize=tiny]{todonotes}

\icmltitlerunning{Normalise for Fairness: A Simple Normalisation Technique for Fairness in Regression Machine Learning Problems}

\begin{document}

\twocolumn[
\icmltitle{Normalise for Fairness: A Simple Normalisation Technique for Fairness in Regression Machine Learning Problems}
% It is OKAY to include author information, even for blind
% submissions: the style file will automatically remove it for you
% unless you've provided the [accepted] option to the icml2023
% package.

% List of affiliations: The first argument should be a (short)
% identifier you will use later to specify author affiliations
% Academic affiliations should list Department, University, City, Region, Country
% Industry affiliations should list Company, City, Region, Country

% You can specify symbols, otherwise they are numbered in order.
% Ideally, you should not use this facility. Affiliations will be numbered
% in order of appearance and this is the preferred way.
\icmlsetsymbol{equal}{*}

\begin{icmlauthorlist}
\icmlauthor{Mostafa M. Amin}{augsburg,tum,syncpilot}
\icmlauthor{Bj\"orn W. Schuller}{augsburg,tum,glam}
\end{icmlauthorlist}

\icmlaffiliation{augsburg}{Chair of Embedded Intelligence for Health Care and Wellbeing, University of Augsburg, Augsburg, Germany}
\icmlaffiliation{tum}{Chair of Health Informatics, Technical University of Munich (TUM), Munich, Germany}
\icmlaffiliation{glam}{Imperial College London, UK}
\icmlaffiliation{syncpilot}{SYNCPILOT GmbH, Augsburg, Germany}

\icmlcorrespondingauthor{Mostafa M. Amin}{mostafa.amin@tum.de}
\icmlcorrespondingauthor{Bj\"orn W. Schuller}{schuller@tum.de}

% You may provide any keywords that you
% find helpful for describing your paper; these are used to populate
% the "keywords" metadata in the PDF but will not be shown in the document
\icmlkeywords{Fairness,Regression,Labelling bias,Normalisation}

\vskip 0.3in
]

% this must go after the closing bracket ] following \twocolumn[ ...

% This command actually creates the footnote in the first column
% listing the affiliations and the copyright notice.
% The command takes one argument, which is text to display at the start of the footnote.
% The \icmlEqualContribution command is standard text for equal contribution.
% Remove it (just {}) if you do not need this facility.

%\printAffiliationsAndNotice{}  % leave blank if no need to mention equal contribution
\printAffiliationsAndNotice{}%\icmlEqualContribution} % otherwise use the standard text.

\begin{abstract}
Algorithms and Machine Learning (ML) are increasingly affecting everyday life and several decision-making processes, where ML has an advantage due to scalability or superior performance.
Fairness in such applications is crucial, where models should not discriminate their results based on race, gender, or other protected groups.
This is especially crucial for models affecting very sensitive topics, like interview invitation or recidivism prediction.
Fairness is not commonly studied for regression problems compared to binary classification problems; hence, we present a simple, yet effective method based on normalisation (\norm), which minimises the impact of unfairness in regression problems, especially due to labelling bias.
We present a theoretical analysis of the method, in addition to an empirical comparison against two standard methods for fairness, namely data balancing and adversarial training.
We also include a hybrid formulation (\normbln), merging the presented method with data balancing, in an attempt to face labelling and sampling biases simultaneously.
The experiments are conducted on the multimodal dataset First Impressions (FI) with various labels, namely Big-Five personality prediction and interview screening score.
The results show the superior performance of diminishing the effects of unfairness better than data balancing, also without deteriorating the performance of the original problem as much as adversarial training.
Fairness is evaluated based on the Equal Accuracy (EA) and Statistical Parity (SP) constraints.
The experiments present a setup that enhances the fairness for several protected variables simultaneously.
%BS: does the suggested method have name? I think it needs one. Perhaps FairNorm? Or SimpleFair? Or FairReg?
%BS: can you name the database by name, please?
\end{abstract}
\section{Introduction}
\label{sec:introduction}

The impact of Algorithms, Artificial Intelligence (AI), and Machine Learning (ML) on our daily lives is increasing day by day, and they became involved in many crucial decision-making processes, \eg loan decisions \cite{loans}, hiring decisions \cite{hiring}, and recidivism \cite{chouldechova2017recidivism}.
ML can be very effective in the automation of tasks that require a lot of manual work, which can have huge cost benefits \cite{brynjolfsson2017can}.
ML can also be effective in areas where data-driven predictions are more reliable than human judgement, this could be due to the ability of algorithms to consider more factors.
For example, \citet{grove2000clinical,meehl1954clinical} discuss a large body of literature that suggest that data and evidence-based assessments can be superior to human judgement in clinical setup.
The use of algorithms and ML in daily life will likely increase in the future, due to an increasing trend of digitalisation in many sections \cite{digitalization}; as a result, the impact of algorithms is likely to increase and affect more sensitive decisions.

A general idea behind \emph{fairness} in ML is training models that do not discriminate their results based on gender, race, or other criteria like those stated in article 2 of the human rights declaration \cite{assembly1948universal}:
\begin{quote}
%\vspace{-0.2cm}
\emph{``Everyone is entitled to all the rights and
freedoms set forth in this Declaration, without
distinction of any kind, such as race, colour, sex,
language, religion, political or other opinion,
national or social origin, property, birth or other
status."}
%\vspace{-0.2cm}
\end{quote}

ML models in general can be unfair in their predictions \wrt\,a \emph{protected} variable (\eg race or gender) by violating fairness constraints; Equal Accuracy (EA) and Statistical Parity (SP) are two commonly adopted fairness constraints \cite{unfairnessML}.
Violating EA can make a model be more accurate for specific values of the protected variable, \eg being more accurate at predicting a variable for males. %females.
Violating SP can make a model predict labels that are biased \wrt\,the protected variable, \eg predicting an interview score for females to be consistently higher than for males.
In these scenarios, the ML models that will be used for decision-making can negatively impact individuals unfairly in a severe manner.
\citet{survey,snapshot} survey many approaches for fairness in ML as well as some of the reasons behind it, and methods to measure it.
Some of the common (not mutually exclusive) reasons for unfairness in ML according to \cite{survey,snapshot} are different data biases, including but not limited to:
\begin{enumerate}[wide, labelindent=0pt, noitemsep,topsep=0pt]
    \item {\bf Labelling bias}, labels can have inherent bias due to their collection mechanism, \eg bias of human annotators.
    \item {\bf Sampling bias}, where there is an imbalance in the sample sizes between different values of the protected variable, in which case maximising the accuracy will prioritise certain groups. %maximising it for the majority class.
    \item {\bf Feature bias}, where some features are more correlated with one (or more) of the protected groups. % than others. 
\end{enumerate}

Collecting more data or using data balancing techniques were studied as ways of encountering sampling bias \cite{balancing}.
Feature bias has been encountered by using adversarial learning \cite{xu2021robust}, where an adversarial model is trained to re-represent the input features in a manner agnostic \wrt\,the protected variable, hence acquiring features that do not leak information about the protected variable, which makes it challenging for the predictor model to discriminate accordingly.
Labelling bias essentially means that the labels have different distributions based on the value of the protected variable, therefore, sampling techniques are solving an orthogonal issue, because a model trained on a properly sampled data with labelling bias will still learn to exploit the unfairness due to labelling bias.
Nevertheless, a poorly sampled data could result in choosing a subset with an apparent labelling bias.

As a motivating example for our method, if we consider creating a model that predicts the height of an American person using their weight as input.
This model would typically attempt to output values around the average height of 170\,cm.
If we modify this model by giving it additional information about gender, it can establish that a male has a height close to the average male height of 177\,cm, while a female's height is close to average female height of 163\,cm
\footnote{Averages obtained on 26.01.202 from: \url{https://www.worlddata.info/average-bodyheight.php}},
thus the model has \emph{unfairly} exploited an advantage based on the implicit hint leaked by gender.
In this example, the information about gender is given explicitly, while in real-world applications this is often only leaked implicitly through the input features, \eg seeing a beard in an input image; this is where unfairness due to feature bias occurs.
Furthermore, height is an `objective' easy-to-measure target variable; however, in other ML applications the labels can be subjective and likely to have an unjustified implicit labelling bias, \eg interview score.

To the best of the authors' knowledge, most of the literature about fairness in ML is concerned with classification problems, while much less are concerned with regression problems.
The few publications concerned with the unfairness in regression problems typically achieve that by quantising the continuous labels, and hence reducing the problem to a classification problem.
Furthermore, to the authors' best knowledge, fairness against labelling bias is also not commonly studied. \cite{blum2019recovering} is the only work we found, addressing labelling bias, but in classification settings.
This is probably because ground truth labels can be hard to acquire, when they require another group of annotators (which can also be biased), or can be hard to define in some cases, like in social signals. %labels (\eg personality predictions).

The contributions of this paper are:
\begin{itemize}[wide, labelindent=0pt, noitemsep,topsep=0pt,leftmargin=2pt]
    \item Introducing a normalisation approach to train fair regression models encountering labelling bias.
    \item Introducing a hybrid approach, merging the presented normalisation approach and data balancing technique, to simultaneously address labelling and sampling biases.
    \item Studying the possibility of protecting from unfairness for two protected variables simultaneously.
\end{itemize}

{
\iffalse
In this paper, we use the multimodal dataset First Impressions (FI) that has videos (images and audio) as input, and personality labels (Big-Five) and an invitation to interview label.
We demonstrate the sampling and labelling biases in the data.
%
In particular, \cite{weisberg2011gender} report that there are differences between gender in some of the Big-Five personality traits, namely females score slightly higher in extroversion, and higher in agreeableness and neuroticism.
A crucial difference in typical personality assessments like in \cite{weisberg2011gender} is that, it is less subject to bias since the outcome is a result of self-answering of a personality questionnaire.
On the other hand, in the FI dataset, the results were not self-answered, but rather estimated by crowdsource annotators which caused a labelling bias in the collected labels \cite{ponce2016chalearn}.
%
Even though the personality could reflect some differences in some protected groups (\eg gender), we believe that the outputs of the models should ideally reflect the \emph{equal opportunity} concept.
%
The reason is, unfairness in a label like the interview could cause a significant harm on an individual, which raises ethical concerns.
%
Additionally, the human decision-makers responsible for deploying the ML model are likely to be enforcing policies that seek to eliminate discrimination in compliance with the human rights act.
\fi
}

The paper is divided as follows:
Related work is discussed in \cref{sec:related}.
In \cref{sec:background}, we discuss the background of the dataset, and fairness methodologies.
We present our method in \cref{sec:approach}, and demonstrate it in experiments in \cref{sec:experiments}.
We conclude with some remarks in \cref{sec:conclusion}.
\section{Related Work}
\label{sec:related}

\citet{Yan2020} has attempted to mitigate bias in the multimodal dataset First Impressions (FI); they used two common approaches, namely data balancing and fairness adversarial learning.
\citet{berk2017convex} introduced convex regularisers that assist linear regression and logistic regression.
\cite{survey,snapshot} are two surveys about fairness in ML, but most approaches are concerned with classification problems, particularly binary classification.
Regression problems did not catch as much attention, however, there are few approaches that mostly rely on quantising the regression labels, hence, transforming the problem into a classification problem, and then use one of several approaches \cite{agarwal2019fair,gorrostieta2019gender}. 
\citet{gorrostieta2019gender} apply this in the scope of Speech Emotion Recognition (SER), while \citet{agarwal2019fair} consider it in crime rate prediction and law school students' GPA prediction.
\citet{narasimhan2020pairwise} provides a framework for pairwise comparisons to optimise fairness constraints in ranking problems.

Due to the bias mischaracterisation in the FI dataset, we believe that \cite{Yan2020} did not actually manage to solve the bias in the FI dataset. 
The reason for this is twofold.

First, they employed adversarial learning which trains a model that learns a different representation of the input features that maintains the performance to be as high as possible whilst not leaking any information about the protected variable.
This technique, however, could severely deteriorate some important features that happen to correlate with the protected variable, for example, an audio feature like pitch correlates with gender  \cite{schuller2013computational}.
The results reported by \cite{Yan2020} for adversarial learning have very poor performance for the original problem, where they score just slightly worse than a constant predictor baseline.
We will demonstrate further why this technique is not capable to address the problem at hand.

Second, \citet{Yan2020} employed data balancing which gives different examples different weights, according to the frequency of the corresponding value of the protected variable.
This technique is usually helpful when there is sampling bias; then, this is conquered by oversampling the less dominant classes, so that all classes have a similar priority for optimisation.
However, when the labels themselves are biased, for example, when females score higher for an interview label, then oversampling the other class (\ie male) will not have an impact on the fact that females are still more likely to be chosen.
Data balancing can be used to downsample some examples, in a manner that makes the labels of the different protected groups similarly distributed; however, this can deteriorate the performance of the original problem since it eliminates some data, especially challenging ones.

\section{Methods and Material}
\label{sec:background}

In this section, we describe the FI dataset\footnote{License: CC BY-NC 4.0\vspace{-0.2cm}}, State-Of-The-Art (SOTA) pipeline (including input features and personality prediction models), fairness baselines, and fairness metrics.

\begin{table}[!t]
    \centering
    \begin{tabular}{c||c|c||c|||c|c||c}
%    & \multicolumn{6}{c}{Gender}\\
%    \hline
    & \multicolumn{3}{c|||}{Train+Dev} & \multicolumn{3}{c}{Test}\\
    \hline
         & M & F & $\Sigma$ & M & F & $\Sigma$\\
         \hline
         Cau.     & 3\,300 & 3\,570 & 6\,870 & 804 & 924  & 1\,728 \\
         \hline
         Asi.      & 82 & 201 &  283 &  14 & 34 &  48 \\
         \hline
         Afr.     & 268 & 579   & 847  & 70 & 154  & 224 \\
         \hline
         \hline
         $\Sigma$ & 3\,650 & 4\,350 & 8\,000 & 888 & 1\,112 & 2\,000 
    \end{tabular}
    \caption{Statistics about the distributions of gender and race in the FI dataset.}
    \vspace{-0.4cm}
    \label{tab:fi-stats}
\end{table}

\subsection{First Impressions Dataset and Personality Prediction Challenge}

The FI dataset was collected for two challenges -- \cite{ponce2016chalearn} and \cite{escalante2017}, where the inputs are 15 seconds videos with one speaker (collected and segmented from YouTube).
In the first challenge \cite{ponce2016chalearn}, the participants were asked to solve the task of predicting the Big-Five personality labels (OCEAN) of the person in the video, namely \emph{Openness to Experience}, \emph{Conscientiousness}, \emph{Extroversion}, \emph{Agreeableness}, and \emph{Neuroticism}.
In the second challenge \cite{escalante2017}, the participants were asked to predict the Big-Five personality dimensions (like the first challenge), in addition to a new label \emph{interview}, which is a score ranking the possibility of an invitation to an interview based on their apparent personality.
The labels of both phases are collected by different sets of annotators, using crowd-sourcing through Amazon Mechanical Turk (AMT), indicating that the labelling bias is coming from independent sources.
The labels are regression values within the range $[0, 1]$.
The distribution of the Test-set labels is detailed in~\cref{fig:dists} (in~\ref{app:dists}).
%Furthermore, in the second challenge, there was a qualitative part, where the participants were asked to provide a system that would give textual recommendation whether a speaker would be invited for an interview, with some reasoning based on the OCEAN features.

Eventually, due to a conceived bias in the predictions, extra metadata was later collected, namely \emph{race} and \emph{gender}.
The gender has two values \emph{male} (M) or \emph{female} (F), and the race has three values \emph{Caucasian} (Cau.), \emph{African-American} (Afr.), and \emph{Asian} (Asi.).
An analysis has shown that there is labelling bias in the ground truth labels, where the females or Caucasians are favoured generally, and African-Americans are unfairly disfavoured for an interview invitation \cite{escalante2020}.
%%% This is shown in \cref{tab:gt-bias}.
%
Even though the correlation values are not very high, they are statistically significant, which indicates %to the fact that there is
a systematic unfairness.
This is further explored by \cite{junior2021person}.

\cite{weisberg2011gender} show that there are differences between genders in some of the personality traits (self-answered), namely, females score slightly higher in extroversion, and higher in agreeableness and neuroticism.
These differences are not the same as in the FI dataset, which indicates that the data has a labelling bias due to human labelling, and not due to actual differences 
in the ground truth.
Furthermore, there is a sampling bias as shown in \cref{tab:fi-stats}. % (in~\ref{app:dists}),
where females and Caucasians are over-represented.

\begin{definition}
{\bf Mean Absolute Accuracy (MAA)}, the performance metric used on the FI dataset, given by:
\vspace{-0.2cm}
\begin{equation}
    \label{eq:MAA}
    1 - \frac{1}{n} \sum_{i=1}^n | y_i - p_i |,
\vspace{-0.2cm}
\end{equation}
where $y_i, p_i$ are the $i^{\text{th}}$ ground truth and predicted examples, respectively; $n$ is the number of evaluation examples.

\end{definition}

% \begin{table}[!t]
%     \centering
%     \begin{tabular}{l||l||l|l|l}
% {Label}   & Female & {Asi.}  & Afr.     & {Cau.}    \\
% \hline                           
%  Agree.  & -0.02       & -0.00 & \bf{-0.07**} & \bf{0.06**} \\
%  Cons.   & \bf{0.08**} & 0.02  & \bf{-0.07**} & \bf{0.06**} \\
%  Extra.  & \bf{0.21**} & 0.04* & \bf{-0.07**} & 0.05*       \\
%  Neuro.  & -0.05*       & 0.00 & 0.05*       & -0.05*       \\
%  Open.   & \bf{0.17**} & 0.01  & \bf{-0.10**} & \bf{0.08**} \\
%  Interv. & \bf{0.07**} & 0.02  & \bf{-0.07**} & 0.05*       \\
%     \end{tabular}
%     \caption{Analysis of the bias in the ground truth labels of the First Impressions dataset, for the Train+Dev portion. The analysis shows Pearson Correlation Coefficient (PCC) between each label and a binary variable of the selected value of a protected variable. The scores for males are not given because they are the negative of the females' scores. The two markers $^{*}, ^{**}$ indicate significance levels with $p$-values $<10^{-3}, 10^{-6}$, respectively.
%     }
%     \label{tab:gt-bias}
%     %\vspace{-0.3cm}
% \end{table}

\subsection{Input Features}

The SOTA pipeline utilised a variety of features \cite{SOTA}, namely facial features, scene, and audio features.
We also utilise the SOTA pipeline and features.
Similar to \cite{SOTA}, we call the facial features as the \emph{face modality}, and the concatenation of the scene and audio features as the \emph{scene modality}.
The protected variable is \emph{not} included explicitly in the input features.
All the features are preprocessed with a MinMaxScaler (trained on the training data) to fit all the input features linearly within the range $[0, 1]$.
A Kernel Extreme Learning Machine (KELM) is used as a regressor; two KELM models were trained for both modalities,
%the face and scene modalities.
their predictions are then stacked using Random Forest (RF) to give the final predictions.

\subsubsection{Face Features}

Faces were extracted from each frame, then aligned (using the Supervised Descent Method (SDM) \cite{Xiong_2013_CVPR}), cropped, and resized to $64\times 64$.
A VGG-Face for emotion recognition was fined-tuned on the FER-2013 dataset (emotions dataset) \cite{representation}; then, the features from the $33^{rd}$ layer were extracted, which produces a descriptor of 4\,096 features for each frame, which is later reduced using five functionals to a static descriptor of 20\,480 features.
The five functionals are mean, standard deviation, slope, offset, and curvature.
Furthermore, Local Gabor Binary Patterns from Three Orthogonal Planes (LGBP-TOP) \cite{Gabor} were used by applying 18 Gabor filters on the aligned facial images, which results in a descriptor with 50\,112 features.
All the facial features are fused together to give $70\,592$ features per video.

\subsubsection{Scene Features}

Scene features were extracted using the VGG-VD-19 network \cite{simonyan2014very} trained for object recognition.
The network was pretrained on the ILSVRC 2012 dataset.
The features are acquired by extracting the output of the $39^{th}$\,layer, yielding 4\,096 features per video.

\subsubsection{Audio Features}

Audio features were extracted using the openSMILE toolkit, choosing the ComParE 2013 acoustic feature set \cite{Schuller13-TI2}.
The features consist of computing 130 low-level descriptor contours (\eg energy, intensity, and FFT spectrum), then reducing them by applying 54 functionals (\eg moments, and LPC autoregressive coefficients) to obtain 6\,373 features per video \cite{Eyben13-RDI,Eyben15-RSA}.

\subsection{Personality Regression Model}

%\subsubsection{Kernel Extreme Learning Machine}

KELM \cite{huang2011extreme} is a method which improves over Extreme Learning Machine (ELM) \cite{ELM}.
%ELM operates similar to a neural network with one hidden layer, where the second layer has a weights matrix $\boldsymbol{\beta} = \mZ^T (\frac{\mI}{C} + \mZ\mZ^T)^{-1} \mY$.
KELM is a kernel formulation of ELM \cite{huang2011extreme}, which showed better results than other models like Support Vector Machines (SVM).
%Faster implementations of KELM are also available \cite{FastKELM}.
KELM operates by constructing a weights matrix $\boldsymbol{\beta}$ that minimises the Mean Squared Error (MSE) between the predictions and the ground truth, $\boldsymbol{\beta}$ is given by:
% \vspace{-0.3cm}
\begin{equation}
\label{eq:kelm}
\boldsymbol{\beta} = (\frac{\mI}{C} + \mK)^{-1} \mY_o,
%\vspace{-0.3cm}
\end{equation}
where $\mK = K(\mX_o, \mX_o)$, $\mX_o$ is the matrix of the input features of the training dataset, $K$ is a kernel function, $C$ is a regularisation scalar parameter, $\mI$ is the identity matrix of size $n\times n$, $n$ is the number of training examples, and $\mY_o$ is a column vector of the training examples of an output label. % (or a matrix of column vectors for multiple labels) of the training dataset.
Accordingly, the prediction for a matrix $\mX$ is
$K(\mX, \mX_o) \cdot \boldsymbol{\beta}.$
%\begin{equation*}
%\vspace{-0.9cm}
%\end{equation*}

Non-linearity can be introduced in the kernel function $K$;
however, 
\citet{SOTA} opt for a linear kernel function, \ie $K(\textbf{A}, \textbf{B}) = \textbf{A}\textbf{B}^T$, in order to reduce overfitting and the number of hyperparameters.
Similar to \cite{SOTA}, we optimised for the regularisation parameter $C$ by using group $k$-fold cross validation, which is using $k$-fold 
%BS: added (please check):
partitioning,  
while ensuring that the sets of speakers between the training sets and the out-of-bag validation sets are disjoint.
We optimise $C$ using Bayesian Optimisation (BO) \cite{BO} for log sampled $C\in[10^{-7}, 10^2]$ and choosing $C$ which jointly maximises the mean accuracy on the hold-out-set for all folds and six labels.
The experiments showed $C$ to be usually close to $10^{-3}$.

The predictions of the two models (best face and best scene models) are stacked using an RF regressor \cite{Breiman2001}.
The number of trees was not specified by \citet{SOTA}, however, we found that $1\,000$ yields similar results.

\subsection{Fairness Baseline Approaches}
\subsubsection{Adversarial Learning}
\label{subsub:adversarial}

Adversarial Learning \cite{xu2021robust} works by training two adversarial models; the first is a predictor model $P$ which predicts the desired label, and the second is a discriminator model $D$ which predicts the protected variable.
This mechanism ensures that the embedding is representative enough for the predictor model to perform well, while not being representative enough for the discriminator to identify the protected variable.
We implement this by having a two-layer (512 units each) filter model $E$ which transforms an input to an embedding. 
The embedding is then used by $P$ to produce the predicted label, and $D$ uses the embedding to predict the value of the protected variable.
Both $P$ and $D$ are single-layer models to avoid indirect leakage.
The models are trained alternatively as shown in~\cref{alg:adversarial} (in~\ref{app:adversarial}).
%\cref{alg:adversarial} in~\ref{app:adversarial} shows the procedure of the training.
There are three hyperparameters, namely the learning rate, and the regularisation parameters $\lambda_1$, and $\lambda_2$ for the two adversarial losses.
The hyperparameters are optimised using BO \cite{BO},
where each is sampled at a log-scale within the range $[10^{-7}, 10^{-2}]$.
We train the models for 20 epochs using the Adam optimisation algorithm \cite{Adam}.
The experiments showed the learning rate to be usually close to $2\cdot 10^{-5}$.

\subsubsection{Data Balancing}
\label{subsub:balance}
Data balancing simply operates by giving different examples different weights $w$.
We give a weight $w_i = \frac{n}{K n_{c_i}}$ for the $i^{th}$ example, where $n_c$ is the number of examples with protected value $c$ for the protected variable, and $K$ is the number of possible values of the protected variable.
Then, we train using a weighted MSE instead of MSE, namely:
\begin{equation}
    \label{eq:balance}
    \mathbb{E}[(\my - \mpr)^2] = \frac{1}{n}\sum_{i=1}^n w_i (y_i - p_i)^2.
    % \vspace{-0.38cm}
\end{equation}

This leads to a modified kernel function for KELM models, $K(\textbf{A}, \textbf{B}) = \textbf{A}\textbf{B}^T \boldsymbol{\Omega}$, where $\boldsymbol{\Omega}$ is a diagonal matrix with the $w_1, \cdots , w_n$ as the diagonal.
This is proved in~\cref{sec:weighted_KELM}.

\subsection{Fairness Constraints and Metrics}
\label{subsec:metrics}

%There are many metrics for fairness in ML.
\citet{survey} discuss several constraints for satisfying fairness in ML. We select the following two:
%
\iffalse
We focus mainly on Statistical Parity (SP) and Equal Accuracy (EA).
%
SP ensures that the labels are statistically independent from the protected variable, that is $\forall a,z \cdot \mathbb{P}(y > z | C=a) = \mathbb{P}(y > z)$. %, for all the protected variable values $a$.
EA ensures that the accuracy of the predictions are independent from the protected variable, that is $\forall a \cdot \mathbb{E}_{C=a} [|\my - \mpr|] = \mathbb{E}[|\my - \mpr|]$, or equivalently
$\forall a,b \cdot \mathbb{E}_{C=a} [|\my - \mpr|] = \mathbb{E}_{C=b}[|\my - \mpr|]$.
$a, b$ are values of the protected variable.
\fi

\begin{definition}
{\bf Statistical Parity (SP)} ensures that the labels are statistically independent from the protected variable $C$, that is $\forall a,z \cdot \mathbb{P}(y > z | C=a) = \mathbb{P}(y > z)$. %, for all the protected variable values $a$.
\end{definition}
\begin{definition}
{\bf Equal Accuracy (EA)} ensures that the accuracy of the predictions are independent from the protected variable $C$, that is $\forall a,b \cdot \mathbb{E}_{C=a} [|\my - \mpr|] = \mathbb{E}_{C=b}[|\my - \mpr|]$. %, or equivalently
%$\forall a,b \cdot \mathbb{E}_{C=a} [|\my - \mpr|] = \mathbb{E}_{C=b}[|\my - \mpr|]$.
\end{definition}

We define three metrics for fairness, that measure to what extent SP and EA are satisfied:

\begin{definition}
{\bf Pearson Correlation Coefficient (PCC)}. PCC measures the correlation between the predictions $p_{1\cdots n}$ and a binary variable corresponding to a specific value $c$ of the protected variable $C$.
PCC will quantify SP by measuring the systematic linear bias of the labels to score consistently higher or lower for a specific value of the protected attribute.
The reasoning behind this is shown in~\ref{app:pcc}.
PCC for the value $c$ is given by:
\begin{equation}
\label{eq:pcc}
    p_c = \frac{\sum_{i=1}^n (\mathbb{I}[C_i = c] - q_c) (p_i - \bar{p})}{\sqrt{\sum_{i=1}^n (\mathbb{I}[C_i = c] - q_c)^2 } \sqrt{\sum_{i=1}^n (p_i - \bar{p})^2}},
    %\vspace{-0.3cm}
\end{equation}
where $q_c$ is the ratio of examples with the value $c$ for the protected variable, and $\bar{p}$ is the average of the predictions.

% This can be checked for statistical significance using $t$-test, by computing the $p$-value:
% \begin{equation}
%     2 \cdot(1 - \Phi[|r_a| \sqrt{\frac{N - 2}{1 - r_a^2}}])
% \end{equation}
\end{definition}

\begin{definition}

{\bf Statistical Parity Metric (SPM)} We define this as the Mutual Information (MI) between the predictions (continuous) with respect to the protected attribute (discrete). It is estimated using the $k$-Nearest Neighbour (kNN) estimation \cite{ross2014mutual}, with $k=3$, as implemented in the library SciPy.
SPM can measure subtle statistical dependencies, \eg in skewed distributions, unlike PCC which measures linear dependencies; however, SPM is harder to interpret.

Unlike \cite{Yan2020}, where the authors computed the difference of MI scores between the true labels and predictions, we evaluate this only for the predictions, because using the difference assumes implicitly that the ground truth labels are not biased, which is not the case to begin with.
\end{definition}

\begin{definition}

{\bf Equal Accuracy Metric (EAM)} Comparison of accuracy (measured by MAA) for different pairs of values $a,b$ for the protected variable, namely:
\vspace{-0.3cm}
\begin{equation}
    \text{EAM}_{a,b} = \cexpected{a}{|\my - \mpr|} - \cexpected{b}{|\my - \mpr|}.
    %\vspace{-0.3cm}
\end{equation}
\end{definition}

\vspace{-0.3cm}

\section{Proposed Approach}
\label{sec:approach}

Given is a dataset $D = \{(\textbf{x}_1, y_1, c_1), \cdots, (\textbf{x}_n, y_n, c_n) \}$, where $\textbf{x}_i$ is the vector of input features, $y_i$ is the output label, and $c_i$ is the value of a protected variable.
%, \eg if the protected variable is \emph{gender}, then $c_i = 1$ for female and $c_i = 0$ for male. 
The labels are assumed to follow the same distribution, with different parameters based on the value of the protected variable, that is $y_i \sim \mathcal{D}(.| \mu_{c_i}, \sigma_{c_i})$.
Before training, we can transform the labels into a corresponding set of \emph{fair} labels $\hat{y}_1, \cdots, \hat{y}_n$, where the model can learn the same distribution, but without leaking information about the protected variable.
The fair labels of the data are given by the transformation:
\vspace{-0.1cm}
\begin{equation}
\label{eq:method}
\hat{y}_i := \frac{y_i - \mu_{c_i}}{\sigma_{c_i}}\sigma + \mu,
    % \vspace{-0.2cm}
\end{equation}
where $\mu_{c_i}, \sigma_{c_i}$ are the mean and standard deviation for all examples with $c_i$ as the value of the protected variable, respectively. $\mu, \sigma$ are the mean and standard deviation of the whole data, respectively.
These are given by:
\vspace{-0.1cm}
%$$\mu_c = \frac{1}{N_c} \sum_{i \in C_c} y_i \text{ and } \sigma_c^2 = \frac{1}{N_c} \sum_{i \in C_c} (y_i - \mu_c)^2$$
\begin{equation}
\label{eq:meanstd}
\begin{aligned}
\mu_c = \mathbb{E}_c&[\my] \text{ and } \sigma_c^2 = \mathbb{V}ar_c[\my] = \mathbb{E}_c [{\my}^2] - \mathbb{E}_c^2[\my], \\
\mu = \mathbb{E}&[\my]  \text{ and  }  \sigma^2 = \mathbb{V}ar[\my] = \mathbb{E}[{\my}^2] - \mathbb{E}^2[\my]. \\
\end{aligned}
    \vspace{-0.3cm}
\end{equation}
%$$\mu_c = \frac{1}{N_c} \sum_{i \in C_c} y_i \text{ and } \sigma_c^2 = \frac{1}{N_c} \sum_{i \in C_c} (y_i - \mu_c)^2$$
%$$\mu = \frac{1}{N} \sum_{i=1}^N y_i \text{ and }\sigma^2 = \frac{1}{N} \sum_{i=1}^N (y_i - \mu)^2$$
\\
To train a model $M$ with parameters $\textbf{W}$, we can simply train it by optimising the loss function $\mathcal{L}$, defined by:
\vspace{-0.2cm}
\begin{equation}
\label{eq:loss}
\mathcal{L}(\my, \mpr) := \mathbb{E}[(\hat{\my} - \mpr)^2] = \frac{1}{n} \sum_{i=1}^n (\hat{y}_i - p_i)^2,
\vspace{-0.2cm}
\end{equation}
where $\mpr = M(\mX; \mW)$ is the model predictions.

%$$ \mathcal{L}(y, p) = \frac{1}{N} \sum_{i=1}^N (\hat{y}_i - M(\textbf{x}_i ; \textbf{W}))^2 + \frac{1}{C} ||\textbf{W}||^2_2$$
% $(\textbf{x} ; \textbf{W})$ 

% The data balancing technique can additionally be incorporated by simply weighting the different examples, which will give the loss function:

% $$ \frac{1}{N} \sum_{i=1}^N \omega_{c_i}(\hat{y}_i - M(\textbf{x}_i ;\textbf{W}))^2 + \frac{1}{C} ||\textbf{W}||^2_2$$

\subsection{Fairness-performance Trade-off}
\begin{theorem}
\label{thm:bigtheorem}
Training a model $M$ with parameters $\textbf{W}$ to minimise the loss function $\mathcal{L}$ (\cref{eq:loss}, which is the MSE after preprocessing the ground truth labels using \cref{eq:method}), is equivalent to minimising the expression:
\begin{equation}
     \text{MSE}(\my, \mpr) + 2\text{ Cov}(\mpr, \my - \hat{\my}),
% \vspace{-0.3cm}
\end{equation}
\end{theorem}
where $\mpr = M(\mX; \mW)$, $\hat{y}_i - y_i$ is a correction term comparing the unfairness in a ground truth label w.\,r.\,t.\ its corresponding fair label, and thus, the covariance term $\text{Cov}(\mpr, \my - \hat{\my})$ corresponds to the \emph{unfairness} of the predicted labels.
\cref{thm:bigtheorem} is proved in~\ref{app:proof}.
%BS: commented out: "are".

%\vspace{-0.5cm}
%BS: Please check - in the pdf appears some box?

% \begin{corollary}
% Optimising $\mathcal{L}$ is equivalent to minimising the bias-variance-unfairness trade-off.
% \end{corollary}

Minimising the covariance term will minimise the MSE term for predictions that are more unfair than the ground truth, that is $p_i < y_i < \hat{y}_i$ or $p_i > y_i > \hat{y}_i$.
However, in all other scenarios, minimising the covariance term will maximise the MSE term, which means that competent models always have a trade-off between fairness and performance, unless the original data has no labelling bias at all (the second term is equal to 0).
%
%Observing the second term, we find that if a model tries to lower, it is impossible for a model to balance both fairness and performance, unless the model has perfect performance or there is no unfairness in the ground truth. %\TODO{revise}
%
Another way to view this is that, the competent models that optimise performance only will always exploit an unfair advantage from the bias in the data.
Trivially, there could be suboptimal models that are worse on both aspects. %fairness and performance.

%{\bf Analysis}
\subsection{Analysis of Theoretical Optimal Scenarios}
\label{subsec:analysis}

In order to analyse how using~\cref{eq:method,eq:loss} affect fairness constraints, we examine the optimal scenario for optimising $\mathcal{L}$, how it affects both SP and EA, and the relevant trade-offs between both fairness constraints.
% yielded by optimising~\cref{eq:method}.
%The following analysis shows that there is a trade-off between optimising for the constraints SP and EA, and that optimising the loss $\mathcal{L}$ given by~\cref{eq:loss} favours SP.
%
\cref{thm:theorem2,thm:corollary2-1,thm:EA} are proved in~\ref{app:analysis}.
In this subsection, the givens are the ground truth $\my$ (with mean $\mu$ and variance $\sigma^2$) and the predictions $\mpr$ (with mean $\bar{p}$ and variance $s^2$), and Pearson correlation coefficient $r$ between $\mpr$ and $\my$.

\begin{theorem}
\label{thm:theorem2}
Optimising MSE leads to $\text{MSE}(\my, \mpr) = \sigma^2(1 - r^2)$, when $\bar{p} = \mu, s =\sigma \cdot r$, and $r$ is maximised.
%Applied for a specific group $c$, this would yield an optimal $\text{MSE}_c(\my, \mpr) = \cvariance{c}{\my}(1 - r^2_c)$, where $ \bar{p}_c = \cexpected{c}{\my}, s_c = \sqrt{\cvariance{c}{\my}} \cdot r_c$, and a maximal $r_c$.
\end{theorem}

\begin{theorem}
    
\label{thm:corollary2-1}
Optimising the loss $\mathcal{L}$ leads to an optimal scenario, where for all values $c$ of the protected variable, $s_c = \sigma r_c, \bar{p} = \bar{p}_c = \mu, \expected{r_c^2}=r^2$, and $r_c$ is maximal.
\end{theorem}

\begin{theorem}

\label{thm:EA}
    %Given are predictions $\mpr$ with a group (for a group $c$) mean $\bar{p}_c$, standard deviation $s_c$, Pearson correlation coefficient $r_c$, and corresponding ground truth $\my$ with group standard deviation $\sigma_c$.
    Optimising for all pairs $a,b$ the difference $|\text{MSE}_a(y, p) - \text{MSE}_b(y, p)|$ will result in an optimal scenario with
    $\bar{p}_c=\mu_c, s_c = \sigma_c \cdot r_c$,
    %$\bar{p}_c=\cexpected{c}{\my}, s_c = \sqrt{\cvariance{c}{\my}} \cdot r_c$,
    while balancing between the values of $r_c$ by satisfying $\variance{\sigma_c^2 (1 - r_c^2)} = 0$.
\end{theorem}

% \iffalse
% %
% First, we use an alternative formulation of MSE \cite{CCC}, namely:
% \begin{equation}
% \text{MSE}(A,B) = \sigma_A^2 + \sigma_B^2 + (\mu_A - \mu_B)^2 - 2\sigma_A \sigma_B r_{AB},
% \end{equation}
% %
% where $\mu, \sigma$ are the mean and standard deviations of the corresponding lists, respectively, and $r$ is the Pearson correlation coefficient.
% \\
% \\
% The loss $\mathcal{L}_a$ for a protected group $a$ can then be given by:
% \begin{equation}
%     \mathcal{L}_a = \sigma^2 + s_a^2 + (\mu - \bar{p}_a)^2 - 2 \sigma s_a r_a ,
% \end{equation}
% %
% where $\bar{p}_a, s_a^2$ are the average and variance of the predictions for the protected group $a$, respectively, while $r_a$ is the Pearson correlation between the predictions and the corresponding ground truth labels for the examples of the protected group $a$.

% If we treat $s_a, \bar{p}_a, r_a$ as independent variables, and solve for their values that yield optimal $\mathcal{L}_a$, we find that $\mathcal{L}_a$ is minimised when $\bar{p}_a = \mu$, $s_a = \sigma r_a$, and $r_a$ is maximised.
% \fi

\subsubsection{Analysis for Statistical Parity}

\label{subsec:sp-analysis}
For SP, we examine the symmetric version of Kullback–Leibler (KL) Divergence between two distributions of two arbitrary values $a, b$ of the protected variable, under the assumption that both are normal distributions, given by:
\vspace{-0.1cm}
\begin{equation}
%\vspace{-0.4cm}
\text{KL}_{a,b} = (\frac{s_a}{s_b})^2 + (\frac{s_b}{s_a})^2 + (\bar{p}_a - \bar{p}_b)^2 (\frac{1}{s_a^2} + \frac{1}{s_b^2}).
% \vspace{-0.45cm}
\end{equation}

KL is minimised when $\bar{p}_a = \bar{p}_b$ and $s_a = s_b$.
According to~\cref{thm:corollary2-1}, minimising $\mathcal{L}_c$ will try to maximise $r_c$ for all $c$.
By assuming that all protected groups are optimised to a similar normalised performance (that is $r_c \approx r^{*}$ for all $c$),
we get that optimising $\mathcal{L}_c$ also optimises KL for all pairs, because
optimally $\bar{p}_a=\bar{p}_b=\mu, s_a=s_b=\sigma r^{*}$,
hence getting closer to satisfying SP.
%Assuming the model has similar normalised performance for all protected groups, that is $r_a \approx r_b \approx r^{*}$, then 
However, KL gets less optimal if there is a higher variance between the normalised performances (that is, higher $\variance{r_c}$ or $\variance{1 - r_c^2}$).
Furthermore, this scenario will minimise MSE for the dataset as a whole, without paying attention to individual groups (unlike optimising MSE or EA), since optimising $\mathcal{L}_c$ will lead to the optimal scenario asserted by~\cref{thm:theorem2}.

\subsubsection{Analysis for Equal Accuracy}

\label{subsec:ea-analysis}

For EA, we analyse a corresponding expression, namely $|\text{MSE}_a(y, p) - \text{MSE}_b(y, p)|$.
According to~\cref{thm:EA},
this can be minimised by minimising
the individual MSEs on each group, while balancing between the different $r_c$ by minimising the expression $\variance{\sigma_c^2(1-r_c^2)}$.
The last constraint is the only distinction to just minimising MSE.
It attempts to balance between the different values of the normalised performances (like SP), however, the balance is weighted by the corresponding variances of the protected groups (unlike SP).
This will, in turn, compromise some of the best performing classes. %, or the ones with higher variances.
However, this is the intended effect of fulfilling EA.
We demonstrate this effect with a constant predictor in~\ref{app:EA_analysis}.
\subsection{Hybrid Approach}

We can construct a hybrid approach between the introduced normalisation technique and the data balancing technique.
Similar to \cref{eq:balance}, we simply adjust the expectation values of $\mu, \sigma, \mathcal{L}$ (defined by \cref{eq:loss,eq:meanstd}) values to be weighted by the weights $w_i = \frac{n}{K n_{c_i}}$.
%
%$$\mu_c = \frac{1}{n_c} \sum_{c_i=c} y_i w_i
%\text{ and } \sigma_c^2 = \frac{1}{n_c}\sum_{c_i=c} w_i (y_i - \mu_c)^2 $$
\vspace{-0.42cm}
\begin{equation}
\label{eq:balnloss}
\begin{aligned}
\mathcal{L}(\my, \mpr) = &\frac{1}{n}\sum_{i=1}^n w_i (\hat{y}_i - p_i)^2, \\ \text{ where } 
\mu = \frac{1}{n} \sum_{i=1}^n y_i w_i &\text{, and }\sigma^2 = \frac{1}{n}\sum_{i=1}^n w_i (y_i - \mu)^2 .
\end{aligned}
%\end{equation}
%\begin{equation}
\vspace{-0.5cm}
\end{equation}
%
%We use the weight values $w_i = \frac{N}{K n_{c_i}}$, where $K$ is the number of values of the protected variable.
%These weights ensure that all examples from the same cluster are having the same weight.
%
%$$\mu = \mathbb{E}[\my] \text{ and }\sigma^2 = \mathbb{V}ar[\my]$$
% \begin{equation*}
% \begin{aligned}
% \mathcal{L}&(\hat{\my}, \mpr) = \frac{1}{N} \sum_{i=1}^N (\hat{y}_i - p_i)^2 = \frac{1}{N} \sum_{i=1}^N (\hat{y}_i - y_i + y_i - p_i)^2 \\
% & = \frac{1}{N} \sum_{i=1}^N [ (\hat{y}_i - y_i)^2 + (y_i - p_i)^2 + 2 (\hat{y}_i - y_i) (y_i - p_i) ] \\
% & =  \text{MSE}(\hat{\my}, \my) + \text{MSE}(\my, \mpr) + 2 \text{ Cov}(\hat{\my} - \my, \my - \mpr)
% \end{aligned}
% \end{equation*}

\begin{figure*}[t!]
    \centering
    % \scalebox{0.92}{
    \includegraphics[width=\textwidth]{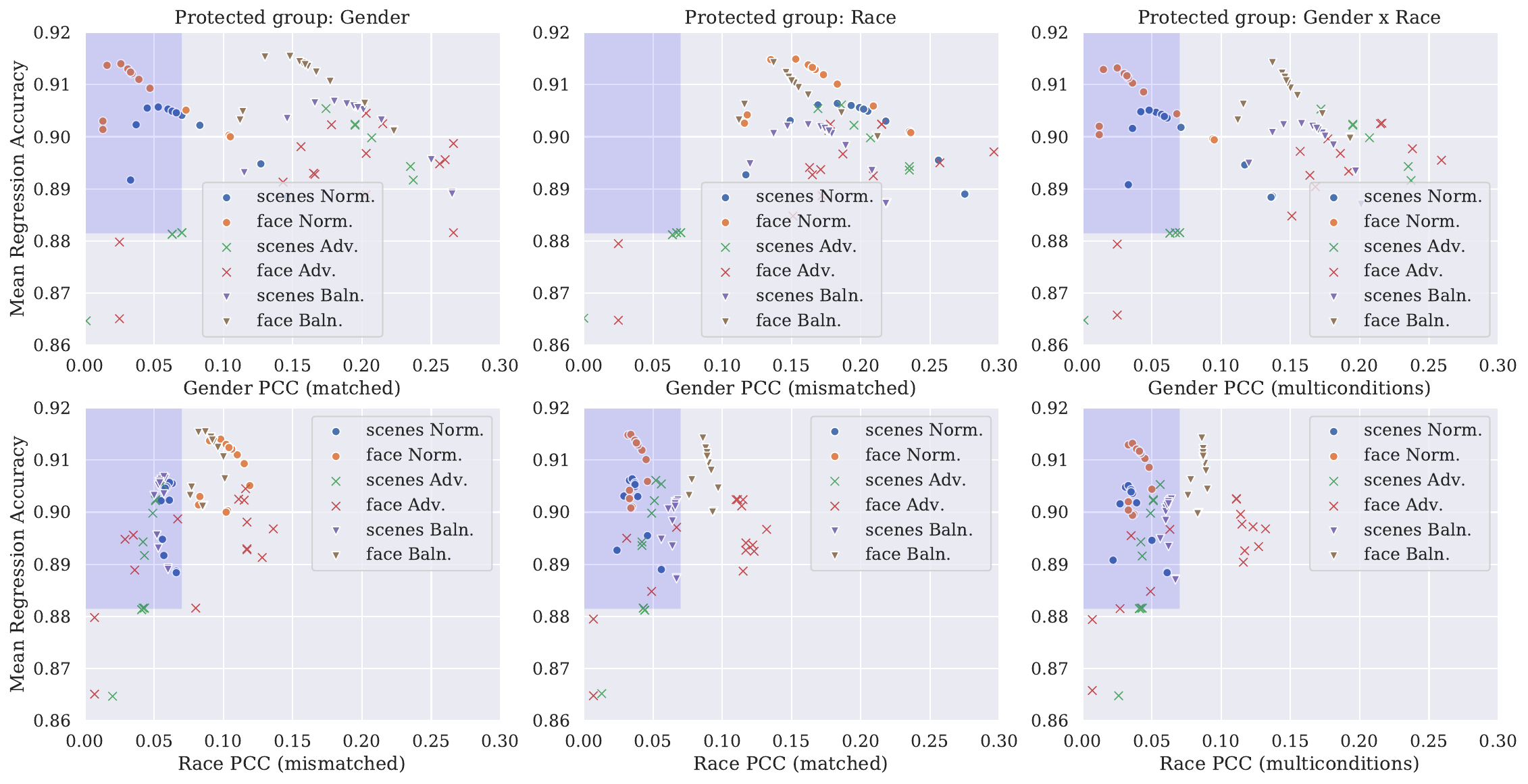}
    % }
    \vspace{-0.4cm}
    \caption{Plotting the relation of MAA and PCC for both gender and race, when the models are trained with gender, race, or both as a protected variable.
    Each point shows the Test-set performance for a hyperparameters configuration sampled by BO during hyperparameters tuning.
    This visualises how difference instances of one method attempts to balance the performance-fairness trade-off.
    The coloured region is where a model has low PCC, with $p$-value not $<10^{-3}$, while maintaining a performance above the constant baseline performance.}
    % \vspace{-0.55cm}
    \label{fig:scatter}
\end{figure*}
\section{Experiments}
\label{sec:experiments}
%In this Section, we present the experimental results, their a discussion, and some limitations. % of the presented methods.

\subsection{Experimental Setup}
\label{subsec:exp-setup}

\begin{table}[!ht]
    \centering
    \vspace{-0.2cm}
    \scalebox{0.9}{
    \begin{tabular}{c|c||c|c|c|c|c|c}
& &    E     &     A     &     C     &     N     &     O     &     I    \\ \hline

\hline \multirow{4}{*}{\rotatebox{90}{Grnd Tr.}}
& F    & {\bf  0.23}  &         0.01  &   \bf  0.09      & \bf     -0.08  & \bf     0.16  &  \bf    0.08 \\
& Asi. &        0.05  &         0.03  &         0.01     &        -0.03  &         0.04  &         0.03 \\
& Cau. & {\bf -0.09}  &  {\bf -0.12}  &  \bf      -0.09  & \bf     0.09  & \bf    -0.13  &  \bf   -0.11 \\
& Afr. &        0.06  &  {\bf  0.09}  &   \bf      0.08  &        -0.07  & \bf     0.10  &  \bf    0.09 \\

\hline \multirow{4}{*}{\rotatebox{90}{Orig.}}
& F    & {\bf  0.26}  &        -0.03  &         0.06  &  {\bf  0.11}  &  {\bf  0.26}  &  {\bf  0.09} \\
& Asi. &        0.02  &         0.01  &        -0.01  &        -0.00  &         0.01  &        -0.00 \\
& Cau. & {\bf  0.10}  &  {\bf  0.13}  &  {\bf  0.12}  &  {\bf  0.11}  &  {\bf  0.13}  &  {\bf  0.12} \\
& Afr. & {\bf -0.12}  &  {\bf -0.15}  &  {\bf -0.13}  &  {\bf -0.12}  &  {\bf -0.14}  &  {\bf -0.13} \\
\hline \multirow{4}{*}{\rotatebox{90}{Gender}}
& F    &        0.06  &         0.00  &        -0.02  &         0.04  &         0.07  &         0.01 \\
& Asi. &        0.01  &         0.01  &        -0.01  &        -0.00  &         0.00  &        -0.00 \\
& Cau. & {\bf  0.12}  &  {\bf  0.12}  &  {\bf  0.13}  &  {\bf  0.12}  &  {\bf  0.15}  &  {\bf  0.13} \\
& Afr. & {\bf -0.14}  &  {\bf -0.13}  &  {\bf -0.13}  &  {\bf -0.13}  &  {\bf -0.17}  &  {\bf -0.13} \\
\hline \multirow{4}{*}{\rotatebox{90}{Race }}
& F    & {\bf  0.27}  &        -0.02  &         0.07  &  {\bf  0.12}  &  {\bf  0.27}  &  {\bf  0.10} \\
& Asi. &        0.01  &         0.01  &        -0.01  &         0.00  &         0.01  &        -0.00 \\
& Cau. &        0.04  &         0.06  &         0.06  &         0.05  &         0.04  &         0.05 \\
& Afr. &       -0.05  &        -0.07  &        -0.06  &        -0.06  &        -0.05  &        -0.06 \\
\hline \multirow{4}{*}{\rotatebox{90}{G $\times$ R}}
& F    &        0.06  &        -0.00  &        -0.02  &         0.04  &         0.06  &         0.01 \\
& Asi. &        0.01  &         0.01  &        -0.02  &        -0.00  &         0.01  &        -0.01 \\
& Cau. &        0.04  &         0.05  &         0.06  &         0.05  &         0.04  &         0.05 \\
& Afr. &       -0.05  &        -0.06  &        -0.05  &        -0.05  &        -0.05  &        -0.05 \\

    \end{tabular}
    }
    %\vspace{-0.2cm}
    \caption{PCC scores for the \norm\ setup with different protected variables at training, showing how a predicted label can leak information about the protected variable. The values in bold highlight when the correlation value is statistically significant with $p$-value $< 0.1\%$ (using $t$-test).
    The values for \normbln\ are very similar to the shown values for \norm, while the values of \bln\ and \adv\ are very similar to the original predictions in the second block.
    }
    \label{tab:PCC_table}
    % \vspace{-0.6cm}
\end{table}

The setup of the experiments aims to test two main factors.
The first is the fairness method,
%for that we have four possibilities,
namely the proposed normalisation technique (\norm), the data balancing technique (\bln), the proposed hybrid approach (\normbln), and adversarial learning (\adv).
The second factor is the protected variable for which the fairness is optimised, namely gender (males or females), race (Caucasian, African-American, or Asian), or a combination of the two (G $\times$ R, six possible combinations).
The \emph{newly introduced} combination of race and gender aims to achieve fairness for both gender and race \emph{simultaneously}. 
% The Gender 
These factors result in a total of twelve setups of four approaches and three protected variables.
%
\iffalse
We categorise the evaluation settings into three categories: (i) \emph{matched settings} (when training and evaluating for the \emph{same} protected variable), (ii) \emph{mismatched settings} (when training and evaluating for \emph{different} protected variables), and (iii) \emph{multi-conditions} (when training is made for all protected variables \emph{jointly}, then evaluating on each protected variable).
\fi
%One advantage of these combinations is that, we can observe the performance of matched (training and evaluating for the matched protected variable), mismatched (training and evaluation for mismatched protected variable), and multi-conditions.

For each of the aforementioned twelve setups, the 8\,000 videos (Train+Dev) are speaker-independently split into 6-folds. %BS: is this reproducible by others? Please provide a URL to paritions if needed.
On each fold, we train a face model and a scene model, where the hyperparameters of the corresponding models are optimised using 
%BS: should this be $BO$ or so?
BO (30 instances per method), by finding the hyperparameters values that yield the best average hold-out-sets performance score (Equations \ref{eq:balance},\ref{eq:loss}, or \ref{eq:balnloss}, depending on the method) for all the six folds and all labels jointly.
This is done separately for the face and scene modalities. %, namely facial features, and merging of scene and audio features.
Eventually, after deciding the best hyperparameters for each of modality, we train one model for each modality on the 8\,000 videos using its corresponding best hyperparameters.
The trained models are then stacked together using RFs,
and we present their results on the Test set (2\,000 videos).

We used an Intel(R) Core(TM) i7-8700 CPU @ 3.20GHz processor, with an Nvidia GeForce RTX 2080 GPU to accelerate the training using~\cref{eq:kelm} for \norm. Training a face model for six folds takes 400\,s on average, while a scene model takes 20\,s.
It takes approximately 75 minutes for all the 30 instances of hyperparameter tuning of \norm\ (both modalities), including an extra overhead time. %, with 15 instances with three protected variables.

We compare the performance of the \emph{original problem}, namely predicting personality and interview invitation, using the MAA metric, on the other hand, we use SPM and PCC for Statistical Parity (SP) fairness assessment, and EAM for Equal Accuracy (EA) fairness assessment.

\subsection{Results}
\label{subsec:results}

The results of the experiments are presented in~\cref{fig:scatter},
%and also in~\ref{app:SP_table}: 
~\cref{tab:PCC_table},
~\cref{tab:SP_table},
and
~\cref{tab:acc_table}
.
%(\cref{tab:SP_table,tab:PCC_table} are in~\ref{app:SP_table}).

\cref{fig:scatter} explores the effects of the performance-fairness trade-off. It shows different instances of the three methods, namely \norm, \bln, and \adv, with both face and scene modalities, for the three protected variables.
Each point is an instance of the corresponding method with different hyperparameters.
The shaded region corresponds to competent models that are achieving well between performance and fairness, by being better than a constant baseline (the mean of the training data), with MAA score $> 0.8815$, whilst not having a high PCC with corresponding $p$-value $<10^{-3}$, checked with a two-tailed $t$-test. %BS: has the significance test method been named? Please do.
Closer to the top-left corner translates to a superior performance in both problem performance (MAA) and SP fairness (PCC).
Outside this region means that the model either has very poor performance, or shows significant leaking of the protected variable. %significant correlation value with the protected variable, which indicates systematic unfairness.
% This region corresponds to the competent models that manage to properly balance between performance and fairness.
The scatter points of each method are resembling a curve, showing that there is a room for improving both performance and fairness simultaneously, but then beyond a certain point, improvement in fairness (PCC) results in a deterioration of the problem performance (MAA).
In~\cref{fig:scatter}, the face-modality models are generally achieving better in both the problem performance and SP fairness, since the scatter points of the face models are generally closer to the top-left corner, compared to their respective scene-modality models.
\bln\ performs very poorly on the SP fairness aspect (most \bln\ models in all figures are outside the competence region), because the unfairness is due to labelling bias, and not sampling bias (even when there is some sampling bias, especially in race).
Furthermore, the models of the \adv\ method are far inferior with far lower MAA, and they achieve SP fairness only when the problem performance (MAA) is worse than a constant baseline.
This is similar to \cite{Yan2020}, where their version of \adv\ reported worse MAA than the constant baseline, which shows that \adv\ method overdoes the anonymaisation of leaking features to the extent of majorly deteriorating the performance of the original problem, which makes this method useless to encounter SP fairness, as it does not solve the original problem at hand.
 
The MAA results of the competent face models are similar to the SOTA results \cite{SOTA}, but with fair predictions (fairness according to SP); as analysed in~\cref{subsec:sp-analysis}.
The performance on the original problem of both \norm, and \bln\ are quite similar as shown in~\cref{tab:acc_table}, and~\cref{fig:scatter}, where they achieve MAA scores close to SOTA \cite{SOTA} results; \normbln\ has slightly worse MAA results, however, \adv\ is achieving much worse MAA results as seen in~\cref{tab:acc_table}. %, especially for the race case, which originally had much higher MAA (Asian).
All methods do not perform well for fairness in the mismatched setups, where the models are trained for fairness on one protected variable an tested on another, which is typical, since they are tested in a situation they are not trained for.

In~\cref{tab:SP_table,tab:PCC_table}, it is clear that \norm\ and \normbln\ are majorly outperforming the two baselines \bln\ and \adv\ in SP fairness, where major information leaks about the protected variable are almost totally eliminated.
%The results of PCC in~\cref{tab:PCC_table} are statistically significant, with a $p$-value $<0.001$ measured with two-tailed $t$-test.
The results of \norm\ and \normbln\ are quite close (especially for PCC), however, \normbln\ is slightly better at SPM metric in~\cref{tab:SP_table} when optimising for race, or gender and race jointly.
This indicates that hybrid approach \normbln\ can account for sampling bias as well as labelling bias, since the effects of sampling bias are more renounced in race (unlike gender). 
\begin{table*}[!ht]%
  \centering
  \subfloat{
    \begin{tabular}{c|c||c|c|c|c|c|c}

\multicolumn{8}{c}{Gender SPM $\times 10$} \\ \hline
& &    E     &     A     &     C     &     N     &     O     &     I    \\ \hline
& Grnd Tr.                  &      .30  &      .04  &      .10  &      .05  &      .14  &      .05 \\
& orig.                     &      .34  &      .00  &      .00  &      .00  &      .59  &      .31 \\ 
\hline \multirow{4}{*}{\rotatebox{90}{Gender}}
& \bln                      &      .26  &   {\bf   .00}  &      .19  &      {\bf .05}  &      .47  &      .34 \\ 
& \norm                     &      .01  &    {\bf  .00}  &    {\bf  .00}  &      .10  &      {\bf .00}  &     {\bf .00} \\ 
& \normbln                  &      {\bf .00}  &     {\bf  .00}  &     {\bf .00}  &      .06  &      {\bf .00}  &      {\bf .00} \\ 
& \adv                      &      .39  &      .07  &      .30  &      .20  &      .47  &      .18 \\ 
\hline \multirow{4}{*}{\rotatebox{90}{Race}}
& \bln                      &      .49  &      {\bf .00}  &     {\bf .00}  &      .19  &    {\bf  .27}  &      .30 \\ 
& \norm                     &      .46  &    {\bf  .00}  &   {\bf .00}  &      .15  &      .29  &    {\bf  .00} \\ 
& \normbln                  &     {\bf .29}  &      .23  &      .08  &  {\bf    .07}  &      .35  &      .10 \\ 
& \adv                      &      .36  &     {\bf .00}  &      .04  &      .18  &      .33  &      .16 \\ 
\hline \multirow{4}{*}{\rotatebox{90}{G $\times$ R}}
& \bln                      &      .42  &      {\bf .00}  &     {\bf .00}  &      .19  &      .37  &      {\bf .00} \\ 
& \norm                     &      .18  &      .05  &   {\bf   .00}  &     {\bf .00}  &      {\bf .00}  &    {\bf  .00} \\ 
& \normbln                  &      {\bf .00}  &     {\bf .00}  &     .18  &      {\bf .00}  &     {\bf .00}  &   {\bf .00}\\ 
& \adv                      &      .75  &      {\bf .00}  &      .03  &      .23  &      .42  &      .10 \\ 
\end{tabular}
  }%
  \qquad
  \subfloat{
    \begin{tabular}{c|c||c|c|c|c|c|c}
\multicolumn{8}{c}{Race SPM $\times 10$} \\ \hline
& &   E     &     A     &     C     &     N     &     O     &     I    \\ \hline
& Grnd Tr.                &      .02  &      .04  &      .13  &      .08  &      .15  &      .08 \\
& orig.                     &      .16  &      .18  &      .19  &      .07  &      .09  &      .25 \\
\hline \multirow{4}{*}{\rotatebox{90}{Gender}}
& \bln                      &      .11  &      .07  &      .22  &      .19  &      .14  &      .17 \\
& \norm                     &      .30  &     {\bf .06}  &      .11  &      .26  &    {\bf  .13}  &     {\bf .08} \\
& \normbln                  &      .14  &      .20  &      {\bf .00} &      {\bf .10}  &      .27  &      .18 \\
& \adv                      &      {\bf .08}  &      .16  &      .03  &      .11  &      .21  &      .08 \\
\hline \multirow{4}{*}{\rotatebox{90}{Race}}
& \bln                     &      .01  &      .14  &      .18  &      {\bf .00}  &      .13  &      .02 \\
& \norm                    &      {\bf .00}  &  {\bf .00}  &      .06  &      .18  &      {\bf .00}  &      .11 \\
& \normbln                 &      .03  &      {\bf .00}  &  {\bf .00}  &      .01  &      {\bf .00}  &      {\bf .00} \\
& \adv                     &      .09  &      .19  &      .12  &      .12  &      .18  &      .06 \\
\hline \multirow{4}{*}{\rotatebox{90}{G $\times$ R}}
& \bln                     &      .16  &      .20  &      .05  &     {\bf .00}  &      .12  &      .10 \\
& \norm                    &    {\bf  .00}  &   .07  &     {\bf .00}  &      .08  &     .10  &   .01 \\
& \normbln                 &    {\bf  .00}  &     {\bf .02}  &      .03  &      .03  &    \bf .03  &   {\bf .00} \\
& \adv                     &      .17  &      .17  &      .23  &      .17  &      .26  &      .13 \\
    \end{tabular}
  }
    \caption{Statistical Parity Metric (SPM) for the four methods, while training for different protected variables (race, gender, or both).}
    \label{tab:SP_table}
\end{table*}
% \end{landscape}

% \begin{landscape}%[t!]
   % \centering
    \begin{table*}
    %\vspace{-.2cm}
    \centering
    \scalebox{0.8}{
    \begin{tabular}{c||c||c|c|c|c||c|c|c|c||c|c|c|c}
    \multicolumn{2}{c||}{} & \multicolumn{4}{c||}{Gender} & \multicolumn{4}{c||}{Race} & \multicolumn{4}{c}{Gender $\times$ Race} \\ 
    \hline
% \multicolumn{14}{c}{} \\
 \multicolumn{14}{c}{Mean Absolute Accuracy $\%$} \\ \hline
    
     &  Orig. & \norm & \bln & \normbln & \adv &   \norm & \bln & \normbln & \adv &    \norm & \bln & \normbln & \adv   \\ \hline
 M     &  91.58  &    {\bf 91.58}  &    91.40  &    91.43  &    91.30  &    91.26  &    {\bf 91.56}  &    91.21  &    91.30  &    91.26  &    {\bf 91.39}  &    91.22  &    91.29 \\
 F     &  91.81  &    {\bf 91.79}  &    91.64  &    91.60  &    91.50  &    91.54  &    {\bf 91.71}  &    91.39  &    91.50  &    {\bf 91.53}  &    91.50  &    91.03  &    91.49 \\
 \hline
 Cau. &  91.70  &    {\bf 91.69}  &    91.50  &    91.50  &    91.42  &    91.37  &    {\bf 91.66}  &    91.27  &    91.42  &    91.35  &    {\bf 91.46}  &    91.06  &    91.41 \\
 Asi.  &  92.62  &    {\bf 92.58}  &    92.26  &    92.29  &    91.91  &    92.64  &    92.63  &    {\bf 92.73}  &    91.86  &   {\bf 92.88}  &    92.25  &    92.41  &    91.93 \\
 Afr.  &  91.58  &    91.56  &    {\bf 91.60}  &    91.56  &    91.24  &    {\bf 91.55}  &    91.34  &    91.32  &    91.23  &    {\bf 91.50}  &    91.21  &    91.23  &    91.21 \\
 
 \hline
% \multicolumn{14}{c}{} \\
 \multicolumn{14}{c}{Equal Accuracy Metric $\%$} \\ \hline
 
F - M  &       0.24  &     0.21  &     0.25  &     {\bf 0.17}  &     0.19  &     0.29  &     {\bf 0.15}  &     0.18  &     0.20  &     0.27  &     {\bf 0.11}  &    -0.19  &     0.20 \\
\hline
\hline
 Af. - C &      -0.12  &    -0.14  &     0.09  &   \bf  0.06  &    -0.18  &     0.18  &    -0.33  &  \bf   0.05  &    -0.19  &   \bf  0.15  &    -0.25  &     0.17  &    -0.20 \\
As. - C &       0.92  &     0.89  &     0.76  &     0.78  &  \bf   0.49  &     1.27  &     0.97  &     1.46  &   \bf  0.44  &     1.53  &     0.79  &     1.35  &  \bf   0.52 \\
As. - Af. & 1.04 & 1.03 & \bf 0.67 & 0.72 & \bf 0.67 & 1.09 & 1.30 & 1.41 & \bf 0.63 & 1.38 & 1.04 & 1.18 &  \bf 0.72 \\
\hline
- & 0.69  &  0.69  &  0.51  &  0.52  &  {\bf 0.45}  &  0.85  &  0.87  &  0.97  &  {\bf 0.42}  &  1.02  &  0.69  &  0.90  &  {\bf 0.48} 
    \end{tabular}
    }
    \vspace{-0.3cm}
    \caption{MAA and EAM scores for the different setups. Bold numbers show the methods with best score for the protected variable. The last row is the mean absolute values of the three race rows.}
    \vspace{-0.5cm}
    \label{tab:acc_table}
    \end{table*}
% \end{landscape}

%  M     &  91.58  &    91.58  &    91.40  &    91.43  &    91.30  &    91.26  &    91.56  &    91.21  &    91.30  &    91.26  &    91.39  &    91.22  &    91.29 \\
%  F     &  91.81  &    91.79  &    91.64  &    91.60  &    91.50  &    91.54  &    91.71  &    91.39  &    91.50  &    91.53  &    91.50  &    91.03  &    91.49 \\
%  \hline
%  Cau. &  91.70  &    91.69  &    91.50  &    91.50  &    91.42  &    91.37  &    91.66  &    91.27  &    91.42  &    91.35  &    91.46  &    91.06  &    91.41 \\
%  Asi.  &  92.62  &    92.58  &    92.26  &    92.29  &    91.91  &    92.64  &    92.63  &    92.73  &    91.86  &    92.88  &    92.25  &    92.41  &    91.93 \\
%  Afr.  &  91.58  &    91.56  &    91.60  &    91.56  &    91.24  &    91.55  &    91.34  &    91.32  &    91.23  &    91.50  &    91.21  &    91.23  &    91.21 \\

The results for EA fairness in~\cref{tab:acc_table} are not as conclusive for the clear superiority of one method achieving EA fairness.
\adv\ is generally performing the best on EAM, as compared to \norm, \normbln, and \bln, especially for race.
For gender, \normbln\ or \bln\ outperform \adv, but \adv\ has reasonable results in this case.
Given the strong performance of \norm\ and \normbln\ for SP fairness, this had the result of compromising the results for EA fairness.
On the other hand, optimising for SP fairness led to a minor compromise in the MAA of the original problem, however, optimising EA led to a bigger compromise in the MAA.
%\todo{explain/analyse why optimising EA has inherently compromising nature with MAA unlike SP.}
These results somewhat agree with the analysis in~\cref{subsec:ea-analysis}, which showed that there is a compromise on accuracy when optimising for EA, and that there are trade-offs when optimising between the two fairness constraints; also, that the proposed method is more suited for SP fairness than EA fairness.

\subsection{Limitations and Potential Negative Impacts}
\label{subsec:limitations}

An important assumption in the provided method is that, all the data are assumed to follow the same family of distributions for each of the protected groups.
However, if this is not the case, for example, if the males' labels follow a gamma distribution, while the females' labels follow a normal distribution, then some advanced models might still encapsulate some information about the protected variable.
We demonstrate this by running a Montecarlo experiment, where we repeat sampling data from a gamma distribution and a normal distribution. In each time, we normalise the data according to \cref{eq:method}, then we measure SPM \wrt\ gender as a binary variable.
By running such an experiment, we find that if the gamma distribution has skewness values $2, \frac{2}{\sqrt{10}}, \frac{2}{10}$, then SPM exhibits the values $0.1, 0.01, 0.006$, respectively.
This shows a scenario where the presented method could get weaker for skewed distributions, which is not the case in the FI dataset (see~\cref{fig:dists} in~\ref{app:dists}).

%There are two potential negative societal impacts.
%First,
Our method gives control over the impact of bias (which we try to neutralise) on training, this can be maliciously used to train a model that systematically discriminates against certain groups, which is a potential negative impact; this can be mitigated by monitoring the fairness metrics in~\cref{subsec:metrics}.
%Second, one must be careful if SP is indeed the proper fairness constraint. In the motivating example in~\cref{sec:introduction}, it could be a requirement in some applications (\eg a clothes shop app) to predict heights suited to the gender; however, doing the same for an interview invitation score is discriminatory.

\section{Conclusion}
\label{sec:conclusion}

%BS: you should start saying you introduced a new method and NAME it :)
In this paper, we introduced a novel method with two variants to mitigate unfairness in regression problems, filling a gap in the fairness literature.
The first is \norm, which focused on eliminating unfairness due to labelling bias.
The second is \normbln, which is a hybrid approach between \norm\ and data balancing; this focused on simultaneously eliminating unfairness due to labelling and sampling biases.
The method consisted of normalising the training labels before training \wrt\ the corresponding value of the protected variable.
We conducted a theoretical analysis showing that, there is always a trade-off between fairness and performance, and between different fairness constraints.
This implied that models that only optimise for performance have to take an unfair advantage of the bias in the data, and that models can not perform the best across all fairness metrics as well as performance.
The experiments confirmed these analyses.

We performed experiments, where we compared the two methods against two alternative methods, namely data balancing and adversarial learning.
The experiments demonstrated that \norm\ and \normbln\ have majorly improved the (Statistical Parity) fairness measures without major deterioration in the performance on the original problem; in comparison, data balancing could not conquer labelling bias, and adversarial learning yielded poor results in the original problem.
Both methods were illustrated to mitigate bias for more than one protected variable simultaneously.
In our experiments, we utilised the First Impressions (FI) dataset,  which is a multimodal dataset, consisting of videos and six regression labels, namely the Big-Five personality features (OCEAN) and a score label for invitation to an interview.
The FI dataset was shown to have labelling and sampling biases, which made it suitable for the presented method.
%We introduced a normalisation technique that can applied as a preprocessing for the labels before training Machine Learning (ML) models.
%The technique is simply normalising all labels with respect to the value of their corresponding protected variable.
%
%We introduced how this technique can be mixed with another fairness technique, namely data balancing.
%
%A discussion was further given on why alternative fairness approaches like adversarial learning and data balancing, which were used in another work, were not suitable for the given problem of labelling bias.

%We also showed how the hybrid approach is specially useful when there are labelling and sampling bias in the data.
%BS: add an OUTLOOK!

% In the unusual situation where you want a paper to appear in the
% references without citing it in the main text, use \nocite
%\nocite{langley00}

\bibliography{references}

\begin{thebibliography}{39}
\providecommand{\natexlab}[1]{#1}
\providecommand{\url}[1]{\texttt{#1}}
\expandafter\ifx\csname urlstyle\endcsname\relax
  \providecommand{\doi}[1]{doi: #1}\else
  \providecommand{\doi}{doi: \begingroup \urlstyle{rm}\Url}\fi

\bibitem[Agarwal et~al.(2019)Agarwal, Dud{\'\i}k, and Wu]{agarwal2019fair}
Agarwal, A., Dud{\'\i}k, M., and Wu, Z.~S.
\newblock {Fair Regression: Quantitative Definitions and Reduction-based
  Algorithms}.
\newblock In \emph{International Conference on Machine Learning}, pp.\
  120--129, Long Beach, California, USA, 2019. PMLR.

\bibitem[Almaev \& Valstar(2013)Almaev and Valstar]{Gabor}
Almaev, T.~R. and Valstar, M.~F.
\newblock {Local Gabor Binary Patterns from Three Orthogonal Planes for
  Automatic Facial Expression Recognition}.
\newblock In \emph{Humaine Association Conference on Affective Computing and
  Intelligent Interaction}, pp.\  356--361, Geneva, Switzerland, 2013. IEEE.

\bibitem[Ambrosio et~al.(2020)Ambrosio, R{\"u}ckert, and Weiss]{digitalization}
Ambrosio, F., R{\"u}ckert, D., and Weiss, C.
\newblock \emph{{Who is prepared for the new digital age? : evidence from the
  EIB investment survey}}.
\newblock European Investment Bank Kirchberg, Luxembourg, 2020.

\bibitem[Berk et~al.(2017)Berk, Heidari, Jabbari, Joseph, Kearns, Morgenstern,
  Neel, and Roth]{berk2017convex}
Berk, R., Heidari, H., Jabbari, S., Joseph, M., Kearns, M., Morgenstern, J.,
  Neel, S., and Roth, A.
\newblock {A Convex Framework for Fair Regression}.
\newblock \emph{arXiv}, 2017.

\bibitem[Blum \& Stangl(2019)Blum and Stangl]{blum2019recovering}
Blum, A. and Stangl, K.
\newblock Recovering from biased data: Can fairness constraints improve
  accuracy?
\newblock \emph{arXiv}, 2019.

\bibitem[Breiman(2001)]{Breiman2001}
Breiman, L.
\newblock {Random Forests}.
\newblock \emph{Machine Learning}, pp.\  5--32, 2001.

\bibitem[Brynjolfsson \& Mitchell(2017)Brynjolfsson and
  Mitchell]{brynjolfsson2017can}
Brynjolfsson, E. and Mitchell, T.
\newblock What can machine learning do? workforce implications.
\newblock \emph{Science}, pp.\  1530--1534, 2017.

\bibitem[Chen et~al.(2018)Chen, Johansson, and Sontag]{balancing}
Chen, I.~Y., Johansson, F.~D., and Sontag, D.
\newblock {Why is My Classifier Discriminatory?}
\newblock In \emph{Proceedings of International Conference on Neural
  Information Processing Systems}, NIPS'18, pp.\  3543--3554, Red Hook, NY,
  USA, 2018. Curran Associates Inc.

\bibitem[Chouldechova(2017)]{chouldechova2017recidivism}
Chouldechova, A.
\newblock {Fair Prediction with Disparate Impact: A Study of Bias in Recidivism
  Prediction Instruments}.
\newblock \emph{Big data}, pp.\  153--163, 2017.

\bibitem[Chouldechova \& Roth(2020)Chouldechova and Roth]{snapshot}
Chouldechova, A. and Roth, A.
\newblock {A Snapshot of the Frontiers of Fairness in Machine Learning}.
\newblock \emph{Communications of the ACM}, pp.\  82--89, 2020.

\bibitem[Cohen et~al.(2019)Cohen, Lipton, and Mansour]{hiring}
Cohen, L., Lipton, Z.~C., and Mansour, Y.
\newblock Efficient candidate screening under multiple tests and implications
  for fairness.
\newblock \emph{arXiv}, 2019.

\bibitem[Escalante et~al.(2017)Escalante, Guyon, Escalera, Jacques, Madadi,
  Baró, Ayache, Viegas, Güçlütürk, Güçlü, van Gerven, and van
  Lier]{escalante2017}
Escalante, H.~J., Guyon, I., Escalera, S., Jacques, J., Madadi, M., Baró, X.,
  Ayache, S., Viegas, E., Güçlütürk, Y., Güçlü, U., van Gerven, M.
  A.~J., and van Lier, R.
\newblock {Design of an Explainable Machine Learning Challenge for Video
  Interviews}.
\newblock In \emph{International Joint Conference on Neural Networks (IJCNN)},
  pp.\  3688--3695, Anchorage, AK, USA, 2017. IEEE.

\bibitem[Escalante et~al.(2020)Escalante, Kaya, Salah, Escalera,
  Güç;lütürk, Güçlü, Baró, Guyon, Jacques, Madadi, Ayache, Viegas,
  Gurpinar, Wicaksana, Liem, Van~Gerven, and Van~Lier]{escalante2020}
Escalante, H.~J., Kaya, H., Salah, A.~A., Escalera, S., Güç;lütürk, Y.,
  Güçlü, U., Baró, X., Guyon, I., Jacques, J. C.~S., Madadi, M., Ayache,
  S., Viegas, E., Gurpinar, F., Wicaksana, A.~S., Liem, C., Van~Gerven, M.
  A.~J., and Van~Lier, R.
\newblock {Modeling, Recognizing, and Explaining Apparent Personality from
  Videos}.
\newblock \emph{Transactions on Affective Computing}, pp.\  1--1, 2020.

\bibitem[Eyben(2015)]{Eyben15-RSA}
Eyben, F.
\newblock \emph{{Real-time Speech and Music Classification by Large Audio
  Feature Space Extraction}}.
\newblock Springer International Publishing, 2015.

\bibitem[Eyben et~al.(2013)Eyben, Weninger, Gro{\ss}, and
  Schuller]{Eyben13-RDI}
Eyben, F., Weninger, F., Gro{\ss}, F., and Schuller, B.
\newblock {Recent Developments in openSMILE, the Munich Open-Source Multimedia
  Feature Extractor}.
\newblock In \emph{Proceedings of ACM International Conference on Multimedia},
  pp.\  835--838, Barcelona, Spain, 2013. Association for Computing Machinery.

\bibitem[Goodfellow et~al.(2013)Goodfellow, Erhan, Carrier, Courville, Mirza,
  Hamner, Cukierski, Tang, Thaler, Lee, Zhou, Ramaiah, Feng, Li, Wang,
  Athanasakis, Shawe-Taylor, Milakov, Park, Ionescu, Popescu, Grozea, Bergstra,
  Xie, Romaszko, Xu, Chuang, and Bengio]{representation}
Goodfellow, I.~J., Erhan, D., Carrier, P.~L., Courville, A., Mirza, M., Hamner,
  B., Cukierski, W., Tang, Y., Thaler, D., Lee, D.-H., Zhou, Y., Ramaiah, C.,
  Feng, F., Li, R., Wang, X., Athanasakis, D., Shawe-Taylor, J., Milakov, M.,
  Park, J., Ionescu, R., Popescu, M., Grozea, C., Bergstra, J., Xie, J.,
  Romaszko, L., Xu, B., Chuang, Z., and Bengio, Y.
\newblock {Challenges in Representation Learning: A Report on Three Machine
  Learning Contests}.
\newblock In \emph{Neural Information Processing}, pp.\  117--124, Berlin,
  Heidelberg, 2013. Springer Berlin Heidelberg.

\bibitem[Gorrostieta et~al.(2019)Gorrostieta, Lotfian, Taylor, Brutti, and
  Kane]{gorrostieta2019gender}
Gorrostieta, C., Lotfian, R., Taylor, K., Brutti, R., and Kane, J.
\newblock {Gender De-Biasing in Speech Emotion Recognition}.
\newblock In \emph{Proceedings INTERSPEECH}, pp.\  2823--2827. ISCA, 2019.

\bibitem[Grove et~al.(2000)Grove, Zald, Lebow, Snitz, and
  Nelson]{grove2000clinical}
Grove, W.~M., Zald, D.~H., Lebow, B.~S., Snitz, B.~E., and Nelson, C.
\newblock {Clinical versus mechanical prediction: a meta-analysis}.
\newblock \emph{Psychological assessment}, pp.\ ~19, 2000.

\bibitem[Huang et~al.(2004)Huang, Zhu, and Siew]{ELM}
Huang, G.-B., Zhu, Q.-Y., and Siew, C.-K.
\newblock {Extreme Learning Machine: A New Learning Scheme of Feedforward
  Neural Networks}.
\newblock In \emph{International Joint Conference on Neural Networks}, pp.\
  985--990, Budapest, Hungary, 2004. IEEE.

\bibitem[Huang et~al.(2011)Huang, Zhou, Ding, and Zhang]{huang2011extreme}
Huang, G.-B., Zhou, H., Ding, X., and Zhang, R.
\newblock {Extreme Learning Machine for Regression and Multiclass
  Classification}.
\newblock \emph{Transactions on Systems, Man, and Cybernetics, Part B
  (Cybernetics)}, pp.\  513--529, 2011.

\bibitem[Junior et~al.(2021)Junior, Lapedriza, Palmero, Bar{\'o}, and
  Escalera]{junior2021person}
Junior, J. C.~J., Lapedriza, A., Palmero, C., Bar{\'o}, X., and Escalera, S.
\newblock {Person Perception Biases Exposed: Revisiting the First Impressions
  Dataset}.
\newblock In \emph{Winter Conference on Applications of Computer Vision
  Workshops (WACVW)}, pp.\  13--21, Waikola, HI, USA, 2021. IEEE.

\bibitem[Kaya et~al.(2017)Kaya, Gurpinar, and Ali~Salah]{SOTA}
Kaya, H., Gurpinar, F., and Ali~Salah, A.
\newblock {Multi-Modal Score Fusion and Decision Trees for Explainable
  Automatic Job Candidate Screening From Video CVs}.
\newblock In \emph{Conference on Computer Vision and Pattern Recognition (CVPR)
  Workshops}, Honolulu, HI, USA, 2017. IEEE.

\bibitem[Kingma \& Ba(2015)Kingma and Ba]{Adam}
Kingma, D.~P. and Ba, J.
\newblock {Adam: A Method for Stochastic Optimization}.
\newblock In \emph{{3rd International Conference on Learning Representations,
  Conference Track Proceedings}}, San Diego, CA, USA, 2015. ICLR.

\bibitem[Meehl(1954)]{meehl1954clinical}
Meehl, P.~E.
\newblock \emph{{Clinical Versus Statistical Prediction: A Theoretical Analysis
  and a Review of the Evidence}}.
\newblock University of Minnesota Press, 1954.

\bibitem[Mehrabi et~al.(2021)Mehrabi, Morstatter, Saxena, Lerman, and
  Galstyan]{survey}
Mehrabi, N., Morstatter, F., Saxena, N., Lerman, K., and Galstyan, A.
\newblock {A Survey on Bias and Fairness in Machine Learning}.
\newblock \emph{ACM Computing Surveys}, 2021.

\bibitem[Mukerjee et~al.(2002)Mukerjee, Biswas, Deb, and Mathur]{loans}
Mukerjee, A., Biswas, R., Deb, K., and Mathur, A.~P.
\newblock {Multi-objective Evolutionary Algorithms for the Risk-return
  Trade-off in Bank Loan Management}.
\newblock \emph{International Transactions in operational research}, pp.\
  583--597, 2002.

\bibitem[Narasimhan et~al.(2020)Narasimhan, Cotter, Gupta, and
  Wang]{narasimhan2020pairwise}
Narasimhan, H., Cotter, A., Gupta, M., and Wang, S.
\newblock {Pairwise fairness for ranking and regression}.
\newblock In \emph{Proceedings of the AAAI Conference on Artificial
  Intelligence}, pp.\  5248--5255, 2020.

\bibitem[Ponce-L{\'o}pez et~al.(2016)Ponce-L{\'o}pez, Chen, Oliu, Corneanu,
  Clap{\'e}s, Guyon, Bar{\'o}, Escalante, and Escalera]{ponce2016chalearn}
Ponce-L{\'o}pez, V., Chen, B., Oliu, M., Corneanu, C., Clap{\'e}s, A., Guyon,
  I., Bar{\'o}, X., Escalante, H.~J., and Escalera, S.
\newblock {Chalearn lap 2016: First Round Challenge on First Impressions -
  Dataset and Results}.
\newblock In \emph{European conference on computer vision}, pp.\  400--418.
  Springer, 2016.

\bibitem[Ross(2014)]{ross2014mutual}
Ross, B.~C.
\newblock {Mutual Information between Discrete and Continuous Data Sets}.
\newblock \emph{PloS one}, pp.\  e87357, 2014.

\bibitem[Schuller \& Batliner(2013)Schuller and
  Batliner]{schuller2013computational}
Schuller, B. and Batliner, A.
\newblock \emph{{Computational Paralinguistics: Emotion, Affect and Personality
  in Speech and Language Processing}}.
\newblock John Wiley \& Sons, 2013.

\bibitem[Schuller et~al.(2013)Schuller, Steidl, Batliner, Vinciarelli, Scherer,
  Ringeval, Chetouani, Weninger, Eyben, Marchi, Mortillaro, Salamin,
  Polychroniou, Valente, and Kim]{Schuller13-TI2}
Schuller, B., Steidl, S., Batliner, A., Vinciarelli, A., Scherer, K., Ringeval,
  F., Chetouani, M., Weninger, F., Eyben, F., Marchi, E., Mortillaro, M.,
  Salamin, H., Polychroniou, A., Valente, F., and Kim, S.
\newblock {The INTERSPEECH 2013 Computational Paralinguistics Challenge: Social
  Signals, Conflict, Emotion, Autism}.
\newblock In \emph{Proceedings INTERSPEECH}, pp.\  148--152, Lyon, France,
  2013. ISCA.

\bibitem[Simonyan \& Zisserman(2015)Simonyan and Zisserman]{simonyan2014very}
Simonyan, K. and Zisserman, A.
\newblock {Very Deep Convolutional Networks for Large-Scale Image Recognition}.
\newblock In \emph{{3rd International Conference on Learning Representations,
  Conference Track Proceedings}}, San Diego, CA, USA, 2015. ICLR.

\bibitem[Snoek et~al.(2012)Snoek, Larochelle, and Adams]{BO}
Snoek, J., Larochelle, H., and Adams, R.~P.
\newblock {Practical Bayesian Optimization of Machine Learning Algorithms}.
\newblock In Pereira, F., Burges, C. J.~C., Bottou, L., and Weinberger, K.~Q.
  (eds.), \emph{Proceedings of International Conference on Neural Information
  Processing Systems}, NIPS'12, pp.\  2951--2959, Red Hook, NY, USA, 2012.
  Curran Associates, Inc.

\bibitem[Tolan et~al.(2019)Tolan, Miron, G\'{o}mez, and Castillo]{unfairnessML}
Tolan, S., Miron, M., G\'{o}mez, E., and Castillo, C.
\newblock {Why Machine Learning May Lead to Unfairness: Evidence from Risk
  Assessment for Juvenile Justice in Catalonia}.
\newblock In \emph{Proceedings of International Conference on Artificial
  Intelligence and Law}, pp.\  83--92, New York, NY, USA, 2019. Association for
  Computing Machinery.

\bibitem[{UN General Assembly}(1948)]{assembly1948universal}
{UN General Assembly}.
\newblock {Universal Declaration of Human Rights}.
\newblock \emph{UN General Assembly}, pp.\  14--25, 1948.

\bibitem[Weisberg et~al.(2011)Weisberg, DeYoung, and Hirsh]{weisberg2011gender}
Weisberg, Y.~J., DeYoung, C.~G., and Hirsh, J.~B.
\newblock {Gender differences in personality across the ten aspects of the Big
  Five}.
\newblock \emph{Frontiers in psychology}, pp.\  178, 2011.

\bibitem[Xiong \& De~la Torre(2013)Xiong and De~la Torre]{Xiong_2013_CVPR}
Xiong, X. and De~la Torre, F.
\newblock Supervised descent method and its applications to face alignment.
\newblock In \emph{Proceedings of Conference on Computer Vision and Pattern
  Recognition (CVPR)}, Portland, OR, USA, 2013. IEEE.

\bibitem[Xu et~al.(2021)Xu, Liu, Li, Jain, and Tang]{xu2021robust}
Xu, H., Liu, X., Li, Y., Jain, A., and Tang, J.
\newblock {To be Robust or to be Fair: Towards Fairness in Adversarial
  Training}.
\newblock In \emph{International Conference on Machine Learning}, pp.\
  11492--11501. PMLR, 2021.

\bibitem[Yan et~al.(2020)Yan, Huang, and Soleymani]{Yan2020}
Yan, S., Huang, D., and Soleymani, M.
\newblock \emph{{Mitigating Biases in Multimodal Personality Assessment}}, pp.\
   361--369.
\newblock Association for Computing Machinery, New York, NY, USA, 2020.

\end{thebibliography}
\bibliographystyle{icml2023}

% \input{sections/checklist}

%%%%%%%%%%%%%%%%%%%%%%%%%%%%%%%%%%%%%%%%%%%%%%%%%%%%%%%%%%%%%%%%%%%%%%%%%%%%%%%
%%%%%%%%%%%%%%%%%%%%%%%%%%%%%%%%%%%%%%%%%%%%%%%%%%%%%%%%%%%%%%%%%%%%%%%%%%%%%%%
% APPENDIX
%%%%%%%%%%%%%%%%%%%%%%%%%%%%%%%%%%%%%%%%%%%%%%%%%%%%%%%%%%%%%%%%%%%%%%%%%%%%%%%
%%%%%%%%%%%%%%%%%%%%%%%%%%%%%%%%%%%%%%%%%%%%%%%%%%%%%%%%%%%%%%%%%%%%%%%%%%%%%%%
\newpage
\onecolumn
\appendix

% \section{Statistical Parity Metric Results}
% \label{app:SP_table}
% The results of the Pearson Correlation Coefficients (PCC) for SP, and the Statistical Parity Metric (SPM) are presented in~\cref{tab:PCC_table,tab:SP_table}, respectively.
% % You can have as much text here as you want. The main body must be at most $8$ pages long.
% % For the final version, one more page can be added.
% % If you want, you can use an appendix like this one, even using the one-column format.
% %%%%%%%%%%%%%%%%%%%%%%%%%%%%%%%%%%%%%%%%%%%%%%%%%%%%%%%%%%%%%%%%%%%%%%%%%%%%%%%
% %%%%%%%%%%%%%%%%%%%%%%%%%%%%%%%%%%%%%%%%%%%%%%%%%%%%%%%%%%%%%%%%%%%%%%%%%%%%%%%

% \input{sections/figures/pearson_table}
% \input{sections/figures/sp_table}

\newpage
\section{Proof of~\cref{thm:bigtheorem}}
\label{app:proof}

\cref{thm:bigtheorem} states that, training a model to minimise the loss function $\mathcal{L}$ (\cref{eq:loss}), is equivalent to minimising the expression:
     $\text{MSE}(\my, \mpr) + 2\text{ Cov}(\mpr, \my - \hat{\my})$.
\begin{proof}
\begin{equation*}
\begin{aligned}
\mathcal{L}&(\hat{\my}, \mpr) = \expected{(\hat{\my} - \mpr)^2} = \expected{(\hat{\my} - \my + \my - \mpr)^2} \\
& = \expected{(\my - \mpr)^2 + (\hat{\my} - \my)^2 + 2 (\hat{\my} - \my)(\my - \mpr)} \\
& = \expected{(\my - \mpr)^2 + (\hat{\my} - \my)^2 + 2 \my(\hat{\my} - \my) + 2 \mpr(\my - \hat{\my}) } \\
& = \expected{(\my - \mpr)^2 + (\hat{\my} - \my)(\hat{\my} + \my) + 2 \mpr(\my - \hat{\my})} \\
& = \expected{(\my - \mpr)^2 + (\hat{\my}^2 - \my^2) + 2 (\mpr - \expected{\mpr})(\my - \hat{\my}) + 2 \expected{\mpr}(\my - \hat{\my}) } \\
& = \expected{(\my - \mpr)^2} + 2 \text{ Cov}(\mpr, \my - \hat{\my}). \\
%= \text{MSE}(\my, \mpr) + 2\text{Cov}(\mpr, \my - \hat{\my}) 
%& =  \text{MSE}(\hat{\my}, \my) + \text{MSE}(\my, \mpr) + 2 \text{ Cov}(\hat{\my} - \my, \my - \mpr)
\end{aligned}
\end{equation*}

The last step is acquired, since by definition
$\expected{\hat{\my}} = \expected{\my}$ and $\mathbb{V}\text{ar}[\hat{\my}] = \mathbb{V}\text{ar}[\my]$, which implies $\expected{\hat{\my} - \my} = \expected{\hat{\my}^2 - \my^2} = 0$.
This equality is the reason why the second and last terms are eliminated.

By minimising both sides w.\,r.\,t.\ the parameters $\mW$, we get:
%
%\begin{aligned}
$$\arg\min_\mW \mathcal{L}(\hat{\my}, \mpr) = \arg\min_\mW  [\expected{(\my - \mpr)^2} + 2 \text{ Cov}(\mpr, \my - \hat{\my})]. $$
%& = \arg\min_\mW  [\mathbb{E}[(\my - \mpr)^2] + 2 (\my - \mpr) (\hat{\my} - \my)] \\
%\end{aligned}

\end{proof}

% \begin{corollary}
% Optimising the loss $\mathcal{L}$ for a protected group $c$ (\cref{eq:loss}) leads to an optimal scenario with $s_c = \sqrt{\variance{\my}} \cdot r_c, \mu_c = \expected{\my}$, and $r_c$ is maximal.
% \end{corollary}

\section{Proofs of the Theorems in~\cref{subsec:analysis}}
\label{app:analysis}

{\bf \cref{thm:theorem2}} asserts that,
given is a ground truth $\my$ (with mean $\mu$ and variance $\sigma^2$) and predictions $\mpr$ (with mean $\bar{p}$, variance $s^2$), and Pearson correlation coefficient $r$ between $\mpr$ and $\my$, then the optimal $\text{MSE}(\my, \mpr) = \variance{\my}(1 - r^2)$, when $\bar{p} = \expected{\my}, s =\sqrt{\variance{\my}} \cdot r$, and $r$ is maximised.
In other words, MSE attempts to get a distribution with the same mean as the original distribution, with maximal similarity between the predictions and original distributions, and with a confidence that is restricted by the normalised performance.
\begin{proof}
    \begin{equation}
        \begin{aligned}
            \text{MSE}(\my, \mpr) &= \expected{(\my - \mpr)^2}
            = \expected{(\my - \expected{\my} - \mpr + \expected{\mpr} + \expected{\my} - \expected{\mpr})^2} \\
            &= \expected{(\my - \expected{\my})^2 + (\mpr - \expected{\mpr})^2 + (\expected{\my} - \expected{\mpr})^2 - \\ &\phantom{xxxxx} 2 (\my - \expected{\my}) (\mpr - \expected{\mpr}) + 2 (\my - \expected{\my} - \mpr + \expected{\mpr})(\expected{\my} - \expected{\mpr}) } \\
            &= \expected{(\my - \expected{\my})^2} + \expected{(\mpr - \expected{\mpr})^2} + \expected{(\expected{\my} - \expected{\mpr})^2} - \\ &\phantom{xxxxx} 2 \expected{(\my - \expected{\my}) (\mpr - \expected{\mpr}) }\\
            &= \variance{\my} + \variance{\mpr} + (\expected{\my} - \expected{\mpr})^2 - 2 \text{ Cov}(\my, \mpr) \\
            &= \sigma^2 + s^2 + (\mu - \bar{p})^2 - 2 \sigma s r
        \end{aligned}
    \end{equation}

The third step is acquired by linearity of expectation, and that fact that $\expected{(\my - \expected{\my} - \mpr + \expected{\mpr})(\expected{\my} - \expected{\mpr})} =0$.

Assuming that $s, \bar{p}, r$ are independent variables, and that $\sigma, \mu$ are constants (corresponding to the ground truth), we can optimise the MSE by differentiating the formula with respect to $s$ and $\bar{p}$.
\begin{equation}
\begin{aligned}
\frac{\partial}{\partial s} \text{MSE}(\my, \mpr) &= 2(s - \sigma r)  \Rightarrow s^{*} = \sigma r \\
\frac{\partial}{\partial \bar{p}} \text{MSE}(\my, \mpr) &= 2 (\bar{p} - \mu) \Rightarrow \bar{p}^{*} = \mu
\end{aligned}
\end{equation}

By substituting these values, we get an optimal MSE of $\sigma^2(1-r^2)$, which is maximised when $r$ gets farther from $0$, ideally when $r = 1$ ($r = -1$ is a rejected solution because $s = \sigma r$, where $\sigma, s \geq 0$).
\end{proof}

{\bf \cref{thm:corollary2-1}} follows directly from~\cref{thm:theorem2}, since $\mathcal{L} = \expected{\mathcal{L}_c}$, and by applying~\cref{thm:theorem2} on each $\mathcal{L}_c$ separately, we get the asserted optimal scenario, which is $\mathbb{E}_c[\mpr] = \expected{\my}$, $\mathbb{V}ar_c[\mpr] = \sigma r_c = \variance{\my} r_c$, and $r_c$ is maximised.
Please note that, in this particular case, $\mathbb{E}_c[\mpr]$ is ideally $\expected{\my}$ and not $\mathbb{E}_c[\my]$, because the loss $\mathcal{L}$ normalises the distributions of $\my$ for the individual protected groups, similarly for $\mathbb{V}ar_c[\mpr]$. The following is a detailed proof:

\begin{proof}

\begin{equation}
    \begin{aligned}
        \mathcal{L}_c &= \cexpected{c}{(\frac{\my - \mu_c}{\sigma_c}\sigma + \mu - \mpr)^2} 
         = \cexpected{c}{(\frac{\my - \mu_c}{\sigma_c}\sigma - (\mpr - \bar{p}_c) + \mu - \bar{p}_c )^2} \\
         &= \cexpected{c}{(\frac{\my - \mu_c}{\sigma_c}\sigma)^2 + (\mpr - \bar{p}_c)^2 + (\mu - \bar{p}_c )^2 - \\ &\phantom{xxxxx} 2 (\frac{\my - \mu_c}{\sigma_c}\sigma)(\mpr - \bar{p}_c) + 2 (\mu - \bar{p}_c)(\frac{\my - \mu_c}{\sigma_c}\sigma - \mpr + \bar{p}_c) } \\
         &= (\frac{\sigma}{\sigma_c})^2\cexpected{c}{(\my - \mu_c)^2} + \cexpected{c}{(\mpr - \bar{p}_c)^2} + (\mu - \bar{p}_c)^2 - 2\frac{\sigma}{\sigma_c}\cexpected{c}{(\my - \mu_c)(\mpr - \bar{p}_c)}\\
         &= (\frac{\sigma}{\sigma_c})^2 \sigma_c^2 + s_c^2 + (\mu - \bar{p}_c)^2 - 2 \frac{\sigma}{\sigma_c} s_c \sigma_c r_c = \sigma^2 + s_c^2 + (\mu - \bar{p}_c)^2 - 2\sigma s_c r_c
    \end{aligned}
\end{equation}

The constraint $sr = \expected{s_c r_c}$ is proved as follows:
\begin{equation}
    \begin{aligned}
         \expected{\mathcal{L}_c} &= \sigma^2 + \expected{s_c^2 + (\mu - \bar{p}_c)^2} - 2\sigma \expected{s_c r_c} \\
         &= \sigma^2 + s^2 + (\mu - \bar{p})^2 - 2 \sigma \expected{s_c r_c} \\
        \mathcal{L} &= \sigma^2 + s^2 + (\mu - \bar{p})^2 - 2 \sigma s r \\
        sr &= \expected{s_c r_c}
    \end{aligned}
\end{equation}

\end{proof}

{\bf \cref{thm:EA}} asserts that minimising $|\text{MSE}_a(\my, \mpr) - \text{MSE}_b(\my, \mpr)|$ (which would satisfy EA) will yield to an optimal scenario where the individual MSE values are minimised, while satisfying the constraint $\variance{\sigma_c^2(1-r_c^2)} = 0$.
To prove that, we consider an arbitrary error function $M_c$ for a specific value $c$ for the protected variable, and an arbitrary dependent variable $v_c$. Then we try to minimise $\mathcal{E} :=  \sum_{a,b} (M_a - M_b)^2$, which will minimise all the aforementioned differences.
\begin{proof}
\begin{equation}
\begin{aligned}
   %\mathcal{E} &=  \sum_{a,b} (M_a - M_b)^2 \\
   \frac{\partial \mathcal{E}}{\partial v_c} &= 2\sum_{a,b}(M_a - M_b)(\frac{\partial M_a}{\partial v_c}  - \frac{\partial M_b}{\partial v_c} )
    = 2\sum_{a,b}(M_a - M_b)(\frac{\partial M_a}{\partial v_a}\delta_a^c  - \frac{\partial M_b}{\partial v_b} \delta_b^c ) \\
    &= 2\sum_{a,b}(M_a - M_b)\frac{\partial M_a}{\partial v_a}\delta_a^c  - 2\sum_{a,b}(M_a - M_b)\frac{\partial M_b}{\partial v_b} \delta_b^c  \\
    &= 4\sum_{a,b}(M_a - M_b)\frac{\partial M_a}{\partial v_a}\delta_a^c = 4 \sum_b (M_c - M_b) \frac{\partial M_c}{\partial v_c} \\
   \end{aligned}
\end{equation}
In the second step, $\frac{\partial M_a}{\partial v_c} = 0$ if $a \neq c$ because $M_a$ isn't depending on $v_c$ by definition, which is reason why assume $\frac{\partial M_a}{\partial v_c} =\frac{\partial M_a}{\partial v_a} \delta_{a}^{c}$. Furthermore, the fourth step is acquired since the two summations are symmetric, we can swap $a, b$ in the second one.

Minimising $\mathcal{E}$ \wrt\,$v_c$ can be achieved when
$\frac{\partial \mathcal{E}}{\partial v_c} = 0$, which implies that $\frac{\partial M_c}{\partial v_c} = 0$ or $\sum_b (M_c - M_b) = 0$ (equivalently $M_c = \expected{M_b}$, or $\variance{M_c} = 0$).
Consequently, minimising $\mathcal{E}$ requires minimising the individual error functions $M_c$ \wrt\, $v_c$ or the variance in $M_c$ is minimal.

Applying this on MSE, we get that $\mathcal{E}$ is minimised when $\text{MAE}_c$ is minimised,
which happens (according to~\cref{thm:theorem2}) when $\bar{p}_c = \mu_c, s_c = \sigma_c r_c$, and $r_c$ is maximal.
In that case, $\text{MAE}_c = \sigma_c^2(1-r_c^2)$. Consequently, $\mathcal{E}$ is minimised when $\variance{\sigma_c^2(1-r_c^2)} = 0$.

\end{proof}

\section{Adversarial Learning Algorithm}
\cref{alg:adversarial} shows the training procedure of one of the baseline fairness methods, namely Adversarial Learning. CE refers to Cross-Entropy, which is the negative log-likelihood.

\label{app:adversarial}
\begin{algorithm}[h!]
   \caption{Adversarial Learning}
   \label{alg:adversarial}
\begin{algorithmic}
    \STATE {\bfseries Input:} Ground truth labels $\my$, protected variable $\textbf{c}$, input features $\mX$
   \STATE {\bfseries Models:} Filter $E$, Predictor $P$, Discriminator $D$
   \FOR{$e$ epochs}
     % \STATE $ \text{MSE}(y, P(E(x))) - \lambda_1 \text{CE}(y, D(E(x))$
     \STATE Train the models $P, E$ one step, while freezing $D$, by minimising: \\
      $\phantom{xxxxxxxx} \text{MSE}(\my, P(E(\mX))) - \lambda_1 \text{CE}(\textbf{c}, D(E(\mX)))$
     \STATE Train the model $D$ one step, while freezing $P, E$, by minimising: \\
     $\phantom{xxxxxxxx} \lambda_2 \text{CE}(\textbf{c}, D(E(\mX)))$
   \ENDFOR
\end{algorithmic}
\end{algorithm}
% \iffalse
% \fi

\section{Proof of Optimal Weighted Kernel Matrix}
\label{sec:weighted_KELM}

Here we prove that~\cref{eq:balance}, corresponds to KELM model with kernel function $K(\textbf{A}, \textbf{B}) = \textbf{A}\textbf{B}^T \boldsymbol{\Omega}$ with optimal Mean Squared Error (MSE) adjusted for data balancing, given by:

\begin{equation*}
    e = \sum_{i=1}^n w_i (y_i - p_i)^2 = \frac{1}{n} (\mY - \mX \boldsymbol{\beta})^T \boldsymbol{\Omega} (\mY - \mX \boldsymbol{\beta}) \\
\end{equation*}

\begin{proof}
The regularised weighted sum of squares error $e$ is given by:
\begin{equation*}
\begin{aligned}
e =& (\mY - \mX \boldsymbol{\beta})^T \boldsymbol{\Omega} (\mY - \mX \boldsymbol{\beta}) + \frac{1}{C} \boldsymbol{\beta}^T\boldsymbol{\beta} \\
e =& (\mY^T\boldsymbol{\Omega}\mY + \boldsymbol{\beta}^T \mX^T \boldsymbol{\Omega} \mX \boldsymbol{\beta} -  \mY^T\boldsymbol{\Omega}  \mX \boldsymbol{\beta} -\boldsymbol{\beta}^T \mX^T \boldsymbol{\Omega} \mY) + \frac{1}{C} \boldsymbol{\beta}^T\boldsymbol{\beta} \\
e =& (\mY^T\boldsymbol{\Omega}\mY + \boldsymbol{\beta}^T \mX^T \boldsymbol{\Omega} \mX \boldsymbol{\beta} - 2 \mY^T\boldsymbol{\Omega}  \mX \boldsymbol{\beta}) + \frac{1}{C} \boldsymbol{\beta}^T\boldsymbol{\beta} \\
\frac{\partial e}{\partial \boldsymbol{\beta}} =& 2 \mX^T\boldsymbol{\Omega}\mX \boldsymbol{\beta} - 2 \mX^T \boldsymbol{\Omega} \mY + \frac{2}{C} \boldsymbol{\beta} \\
\end{aligned}
\end{equation*}
In order to minimise $e$, we solve for $\boldsymbol{\beta}$ for $\frac{\partial e}{\partial \boldsymbol{\beta}} = \textbf{0}$, which implies:
\begin{equation*}
 (\mX^T\boldsymbol{\Omega}\mX + \frac{1}{C} \mI)\boldsymbol{\beta} = \mX^T \boldsymbol{\Omega} \mY
\end{equation*}
There are two solutions for $\boldsymbol{\beta}$, namely 
$\boldsymbol{\beta} = (\mX^T\boldsymbol{\Omega}\mX + \frac{1}{C} \mI)^{-1}\mX^T \boldsymbol{\Omega} \mY$, or $\mX^T \boldsymbol{\Omega} (\mX \mX^T\boldsymbol{\Omega} + \frac{1}{C} \mI)^{-1} \mY$, both can be verified by the substitution in the last equation.
The second solution gives the adjusted kernel function in \cref{subsub:balance}.
\end{proof}

\section{First Impressions Dataset Distributions}
\label{app:dists}
% \input{sections/figures/fi_stats}

\begin{figure*}[!ht]
    \centering
    \includegraphics[width=\textwidth]{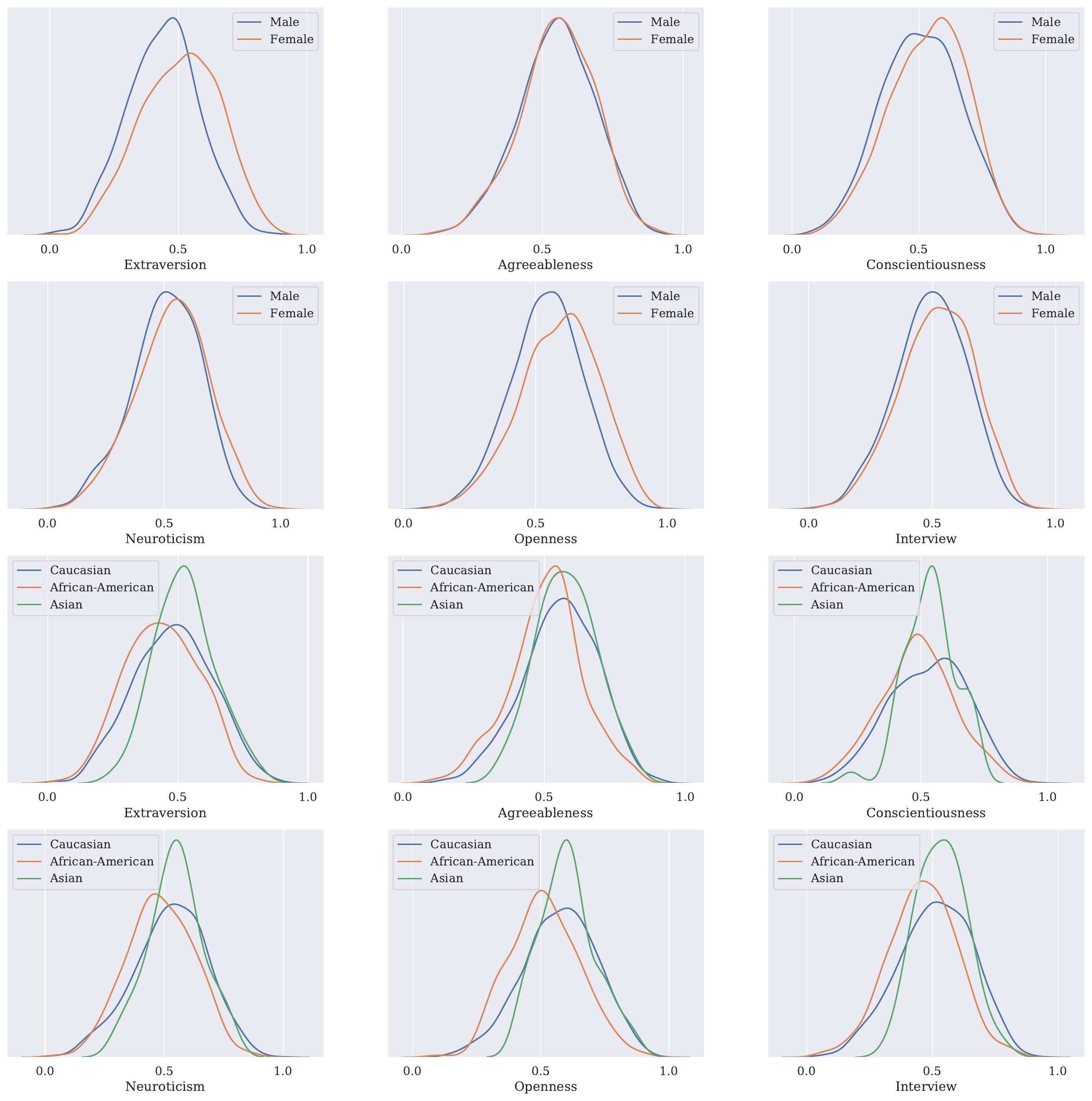}
    \caption{The distributions of the Test-set ground truth labels for the six labels, and each protected variable.}
    \label{fig:dists}
\end{figure*}

% \cref{tab:fi-stats} shows the sizes of different partitions of the FI dataset.
\cref{fig:dists} demonstrates the distributions of the labels for all genders and all races.

\iffalse
\begin{table}[!t]
    \centering
    \begin{tabular}{c||c|c||c|||c|c||c}
%    & \multicolumn{6}{c}{Gender}\\
%    \hline
    & \multicolumn{3}{c|||}{Train+Dev} & \multicolumn{3}{c}{Test}\\
    \hline
         & M & F & $\Sigma$ & M & F & $\Sigma$\\
         \hline
         Cau.     & 3\,300 & 3\,570 & 6\,870 & 804 & 924  & 1\,728 \\
         \hline
         Asi.      & 82 & 201 &  283 &  14 & 34 &  48 \\
         \hline
         Afr.     & 268 & 579   & 847  & 70 & 154  & 224 \\
         \hline
         \hline
         $\Sigma$ & 3\,650 & 4\,350 & 8\,000 & 888 & 1\,112 & 2\,000 
    \end{tabular}
    %\vspace{-0.2cm}
    \caption{Statistics about the distributions of gender and race in the FI dataset.}
    \label{tab:fi-stats}
\end{table}
\fi

\section{Statistical Parity and Pearson Correlation Coefficient}
\label{app:pcc}
Here we prove that Pearson Correlation Coefficient (PCC) defined by~\cref{eq:pcc} corresponds to a quantification measuring Statistical Parity (SP) constraint.

\begin{proof}
The definition of SP:
\begin{equation*}
    \forall z \cdot \mathbb{P}(p > z | C = c) = \mathbb{P}(p > z)
\end{equation*}
By expanding the definition, we get:
\begin{equation*}
    \forall z \cdot \frac{\sum_{i=1}^n \mathbb{I}[p_i > z] \mathbb{I}[C = c_i]}{\sum_{i=1}^n \mathbb{I}[C = c_i]}  = \frac{1}{n} \sum_{i=1}^n \mathbb{I}[p_i > z]
\end{equation*}

\begin{equation*}
\forall z \cdot \frac{1}{n}\sum_{i=1}^n \mathbb{I}[p_i > z] \mathbb{I}[C = c_i] - \frac{1}{n}\sum_{i=1}^n \mathbb{I}[C = c_i] \frac{1}{n} \sum_{i=1}^n \mathbb{I}[p_i > z] = 0
\end{equation*}

By integrating both sides \wrt\ d$z$ for all $z\in[0,1]$, we get:

\begin{equation*}
    \frac{1}{n}\sum_{i=1}^n p_i \mathbb{I}[C=c_i] - \frac{1}{n}\sum_{i=1}^n \mathbb{I}[C=c_i] \frac{1}{n} \sum_{i=1}^n p_i = 0
\end{equation*}

\begin{equation*}
    \frac{1}{n}\sum_{i=1}^n p_i \mathbb{I}[C=c_i] - n_c \bar{p} = 0
\end{equation*}
\begin{equation*}
    \frac{1}{n}\sum_{i=1}^n (\mathbb{I}[C=c_i] - n_c)(p_i - \bar{p}) = 0
\end{equation*}

The left hand side of the equation is the covariance computed by the numerator of PCC, which is equal to $0$ if SP is satisfied.
The further this value from $0$, the more dependent both variable are, hence further away from SP.
Therefore, the absolute value of PCC is a normalised measure that indicates how far SP is satisfied.

\end{proof}

\section{Demonstration for Equal Accuracy Optimisation}
\label{app:EA_analysis}
In this section, we construct a constant predictor that predicts a value $p^*$, which minimises a measure corresponding to the Equal Accuracy constraint.

For a normally distributed labels with a mean $\mu$ and variance $\sigma^2$, the expected mean squared error for a constant prediction $p^*$ is given by, $\mathbb{E}[(p - y)^2] = \int_{-\infty}^\infty (p - y)^2 \phi(y | \mu, \sigma^2) \text{ d}y = (p - \mu)^2 + \sigma^2$. Hence, we present a smoothed version of the Equal Accuracy Metric, which we call Squared Equal Accuracy Metric (SEAM), this is given by:
\begin{equation*}
\begin{aligned}
E &:= \sum_{a,b} (\mathbb{E}_a[(p - y)^2] - \mathbb{E}_b[(p - y)^2])^2\\
  &= \sum_{a,b} [(p - \mu_a)^2 - (p - \mu_b)^2+ \sigma_a^2 -\sigma_b^2]^2 \\
\end{aligned}
\end{equation*}

In order to find the optimal point $p^*$, with minimum value of SEAM, we need to solve $\frac{\partial E}{\partial p} = 0$.
\begin{equation*}
\begin{aligned}
%E &= \frac{1}{m(m-1)}\sum_{a,b} (\mathbb{E}_a[(p - y)^2] - \mathbb{E}_b[(p - y)^2])^2\\
\frac{\partial E}{\partial p} &= 4 \sum_{a, b} ((p - \mu_a)^2 - (p - \mu_b)^2+ \sigma_a^2 -\sigma_b^2) (\mu_b - \mu_a) \\
&= 4 \sum_{a, b} (\mu_a^2 - \mu_b^2 -2(\mu_a - \mu_b) p + \sigma_a^2 -\sigma_b^2) (\mu_b - \mu_a)
\end{aligned}
\end{equation*}
Solving for $\frac{\partial E}{\partial p} = 0$, yields the following:

\begin{equation*}
\begin{aligned}
p^{*} &= \frac{\sum_{a,b} (\mu_a^2 - \mu_b^2 + \sigma_a^2 - \sigma_b^2)(\mu_a - \mu_b)}{2\sum_{a,b} (\mu_a - \mu_b)^2} \\ &= \frac{\sum_{a,b} (\mu_a-\mu_b)^2(\mu_a + \mu_b)/2}{\sum_{a,b}(\mu_a - \mu_b)^2} + \frac{\sum_{a,b} (\sigma_a^2 - \sigma_b^2)(\mu_a - \mu_b)}{2\sum_{a,b} (\mu_a - \mu_b)^2}
\end{aligned}
\end{equation*}

This formula has two components, the first is given by the weighted average between the midpoints of each pair of the protected groups.
The weights are given by the intra-cluster square distances of the corresponding pairs.
This component is independent from the variance of the data.
The second component of the formula is a correction based on the variance, by shifting the ideal constant prediction towards clusters with higher variance.

\begin{figure*}[t!]
    \centering
    \includegraphics[width=\textwidth]{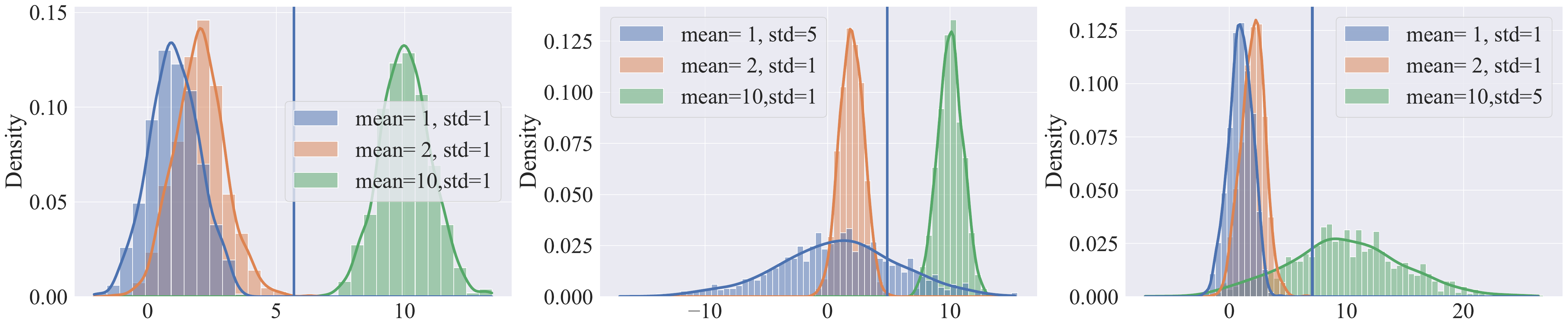}
    \caption{Demonstration of how the constant optimal point $p^*$ (yielding optimal Equal Accuracy) changes based on the variances of the distributions of the protected groups.
    $p^*$ is demonstrated by the vertical line.}
    \label{fig:equalacc-demo}
\end{figure*}

The optimal $p^{*}$ is demonstrated in~\cref{fig:equalacc-demo}, the demonstration has three clusters with means 1, 2, and 10 respectively. In the first figure, all distributions have the same standard deviation; the optimal point $p^{*}$ is near 5.7, this is the value of the first component of the formula in all the three figures.
The optimal point is tilted towards the third distribution since it is much further from the other two. The other two figures show similar analysis, but by changing the standard deviation of the left-most and right-most distributions, respectively.
It can be seen that the optimal point $p^{*}$ moves closer to the distribution with higher variance, by a shift of roughly 2.9 towards the corresponding distributions.

\end{document}